\documentclass[letterpaper]{article} 
\usepackage{aaai25}  
\usepackage{times}  
\usepackage{helvet}  
\usepackage{courier}  
\usepackage[hyphens]{url}  
\usepackage{graphicx} 
\usepackage{natbib}  
\usepackage{caption} 
\usepackage{algorithm}
\usepackage{algorithmic}
\usepackage{multirow}
\usepackage{hhline}
\usepackage{pifont}
\usepackage{booktabs}
\usepackage{amsmath}
\usepackage{amssymb}
\usepackage{graphicx}
\usepackage{xcolor}
\usepackage{enumitem}
\usepackage{colortbl}

\usepackage{newfloat}
\usepackage{listings}
\DeclareCaptionStyle{ruled}{labelfont=normalfont,labelsep=colon,strut=off} 
\floatstyle{ruled}
\newfloat{listing}{tb}{lst}{}
\floatname{listing}{Listing}
\title{DeMo: Decoupled Feature-Based Mixture of Experts for Multi-Modal Object Re-Identification}
\author{
    Yuhao Wang\textsuperscript{\rm 1},
    Yang Liu\textsuperscript{\rm 1,3},
    Aihua Zheng\textsuperscript{\rm 2,3},
    Pingping Zhang\textsuperscript{\rm 1,3}\thanks{Corresponding author (zhpp@dlut.edu.cn).}
}
\affiliations{
    \textsuperscript{\rm 1}School of Future Technology, School of Artificial Intelligence, Dalian University of Technology\\
    \textsuperscript{\rm 2}School of Artificial Intelligence, Anhui University\\
    \textsuperscript{\rm 3}Anhui Provincial Key Laboratory of Multimodal Cognitive Computation, Anhui University\\

    924973292@mail.dlut.edu.cn, \{ly, zhpp\}@dlut.edu.cn, ahzheng214@foxmail.com
}

\usepackage{bibentry}

\begin{document}

\maketitle

\begin{abstract}
    Multi-modal object Re-IDentification (ReID) aims to retrieve specific objects by combining complementary information from multiple modalities.
    Existing multi-modal object ReID methods primarily focus on the fusion of heterogeneous features.
    However, they often overlook the dynamic quality changes in multi-modal imaging.
    In addition, the shared information between different modalities can weaken modality-specific information.
    To address these issues, we propose a novel feature learning framework called DeMo for multi-modal object ReID, which adaptively balances decoupled features using a mixture of experts.
    To be specific, we first deploy a Patch-Integrated Feature Extractor (PIFE) to extract multi-granularity and multi-modal features.
    Then, we introduce a Hierarchical Decoupling Module (HDM) to decouple multi-modal features into non-overlapping forms, preserving the modality uniqueness and increasing the feature diversity.
    Finally, we propose an Attention-Triggered Mixture of Experts (ATMoE), which replaces traditional gating with dynamic attention weights derived from decoupled features.
    %
    %
    With these modules, our DeMo can generate more robust multi-modal features.
    Extensive experiments on three multi-modal object ReID benchmarks fully verify the effectiveness of our methods.
    The source code is available at https://github.com/924973292/DeMo.
\end{abstract}

\begin{figure}[t]
    \centering
    \includegraphics[width=0.4\textwidth]{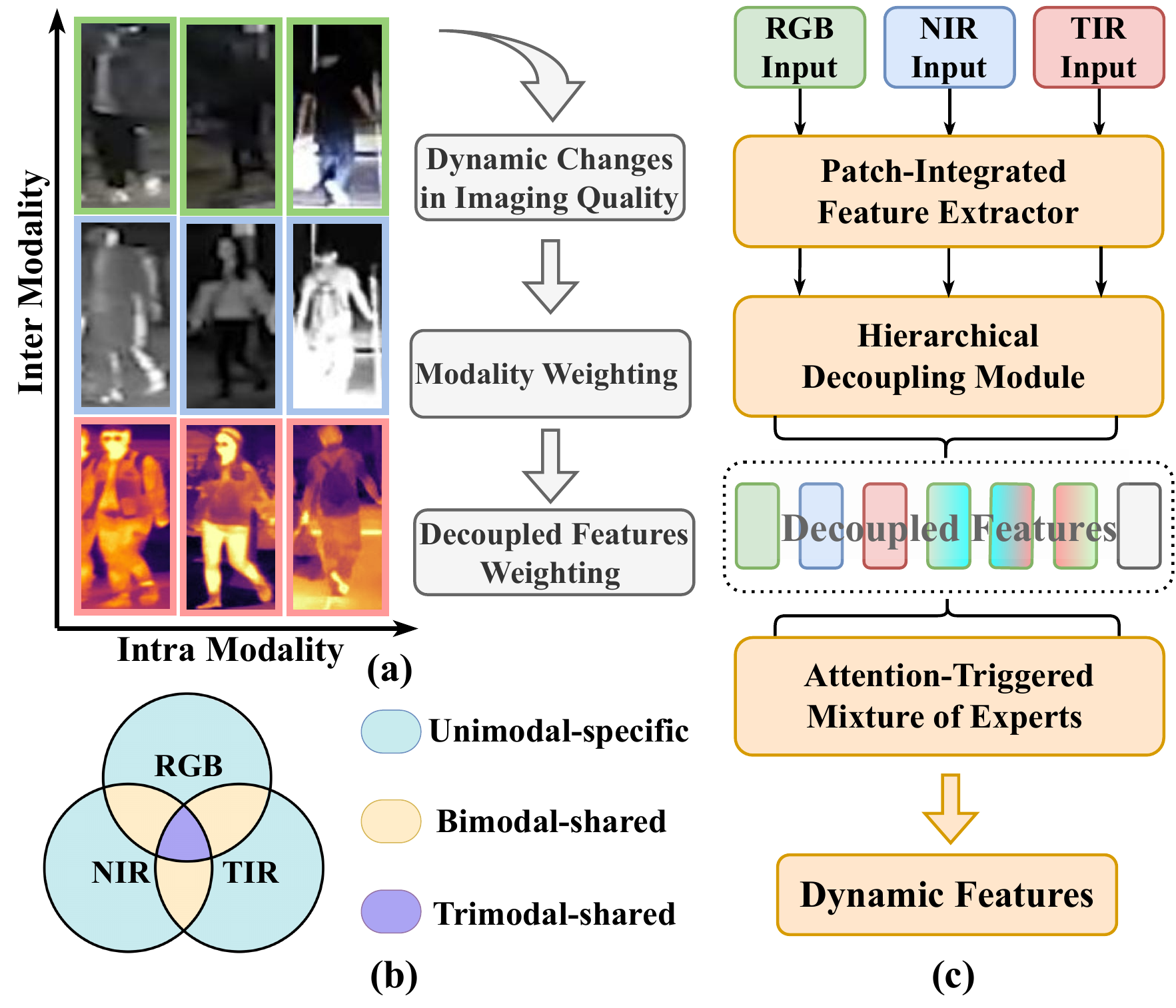}
    \vspace{-4mm}
    \caption{
    (a) The prevalent dynamic quality changes in multi-modal imaging.
    (b) Hierarchical feature decoupling.
    (c) The proposed modules and the framework of our DeMo.}
    \label{fig:Introduction}
    \vspace{-4mm}
    \end{figure}
\section{Introduction}
Object Re-IDentification (ReID) aims to retrieve the same object across different camera views.
Over the past decade, single-modal object ReID~\cite{liu2023deeply,wang2024other,liu2024video,liu2021watching,zhang2021hat,yu2024tf}, primarily based on RGB images, has made significant progress.
However, RGB imaging is highly susceptible to adverse conditions such as darkness and glare, leading to poor generalization in complex environments.
Fortunately, multi-modal imaging, which introduces diverse information from different modalities~\cite{chen2024visible,shi2024multi,shi2023dual,shi2024learning,zheng2021robust,lu2023learning}, has emerged as a promising solution to enhance the feature robustness in challenging scenarios.
By integrating complementary information from different modalities, existing multi-modal object ReID methods~\cite{zhang2024magic,yang2024shallow,Yang_2023_ICCV} achieve remarkable performance.
However, they often overlook dynamic quality changes inherent in multi-modal imaging.
As shown in Fig.~\ref{fig:Introduction} (a), images from RGB, Near Infrared (NIR) and Thermal Infrared (TIR) modalities are presented, respectively.
Horizontally, environmental and imaging interferences lead to content fluctuations within the same modality.
Vertically, the relevance of each modality varies under identical imaging conditions.
For example, in the second column, the RGB image provides limited information due to darkness, whereas the NIR and TIR images clearly show details such as glasses and bags.
Consequently, the model should prioritize the NIR and TIR modalities in such scenarios.
This highlights the need for adaptive modality weighting to address the dynamic changes in multi-modal imaging quality.

Recently, Mixture of Experts (MoE)~\cite{cai2024survey} have gained attention for their effectiveness in adaptively weighting expert features.
Inspired by this, we treat each modality as an expert and dynamically adjust the weights of the experts based on the importance of each modality.
However, directly weighting multi-modal features risks modality information confusion~\cite{zhang2024multimodal}.
As depicted in Fig.~\ref{fig:Introduction} (b), multi-modal features can be categorized into three types: modality-specific, bimodal-shared and trimodal-shared features.
Directly weighting features from RGB, NIR and TIR modalities can amplify modality-shared information while suppressing modality-specific information.
%
Fortunately, decoupling multi-modal features can effectively separates modality-specific information~\cite{wei2024robust}, preserving discriminative details.
Motivated by these observations, we propose DeMo, a novel feature learning framework that adaptively assigns weights to decoupled features with a mixture of experts for multi-modal object ReID.

As shown in Fig.~\ref{fig:Introduction} (c), our DeMo consists of three components: Patch-Integrated Feature Extractor (PIFE), Hierarchical Decoupling Module (HDM) and Attention-Triggered Mixture of Experts (ATMoE).
We first utilize the PIFE to extract multi-granularity representations from multi-modal inputs.
Specifically, the PIFE integrates high-level patch tokens with class tokens to generate robust features.
This synergy between global and local information significantly enhances the feature discrimination for each modality.
Then, we introduce the HDM to guide the hierarchical decoupling of multi-modal features.
We first categorize these features into three hierarchical types based on their degree of information overlap.
By utilizing learnable queries with different combinations of multi-modal tokens, HDM employs cross-attentions to effectively decouple features into corresponding hierarchical levels, preserving modality-specific information and enhancing feature diversity.
Finally, we introduce the ATMoE to replace traditional gating with attention-guided interactions between decoupled features, allowing for more accurate and context-aware weighting of each expert.
Moreover, the multi-head mechanism in ATMoE enhances the model's adaptability to dynamic imaging conditions.
With the above modules, our DeMo can extract robust representations across various scenarios.
Even in extreme cases where one or more modalities are missing, our DeMo can still achieve competitive performances.
Extensive experiments on three multi-modal object ReID datasets validate the effectiveness of our proposed method.

In summary, our contributions are as follows:
\begin{itemize}
    \item
    We introduce DeMo, a novel framework for multi-modal object ReID.
    To our best knowledge, our proposed DeMo is the first attempt to address dynamic changes in multi-modal imaging with decoupled feature-based MoE.
    \item
    We develop a Hierarchical Decoupling Module (HDM) to effectively decouple multi-modal features into hierarchical types, increasing the decoupled features' diversity.
    \item
    We propose an Attention-Triggered Mixture of Experts (ATMoE), which utilizes attention-guided interactions for accurate expert weighting and a multi-head mechanism for adaptability to dynamic imaging conditions.
    \item
    Extensive experiments on three multi-modal object ReID datasets demonstrate the effectiveness of our method.
\end{itemize}
\section{Related Work}
\subsection{Multi-Modal Object Re-Identification}
Multi-modal object ReID has attracted increasing attention due to its robustness in practical scenarios.
Existing methods primarily focus on integrating complementary information from different modalities.
For multi-modal person ReID, Zheng \emph{et al.}~\cite{zheng2021robust} propose to learn robust features with a progressive fusion.
Wang \emph{et al.}~\cite{wang2022interact} introduce an interact-embed-enlarge framework to boost the modality-specific knowledge.
In addition, Zheng \emph{et al.}~\cite{zheng2023dynamic} address the modal-missing problem with a pixel reconstruction method.
%
For multi-modal vehicle ReID, Li \emph{et al.}~\cite{li2020multi} propose to fuse multi-modal features with a coherence loss.
Afterwards, many CNN-based methods~\cite{he2023graph,guo2022generative,zheng2022multi} have been proposed to enhance the feature robustness with modality generation, graph learning and instance sampling, etc.
With the strong generalization ability of vision Transformer~\cite{dosovitskiy2020image} (ViT), many Transformer-based methods~\cite{pan2023progressively,crawford2023unicat,wang2023top,wang2024heterogeneous,zhang2024magic} have been proposed to further improve the performance of multi-modal object ReID.
Among them, Wang \emph{et al.}~\cite{wang2024heterogeneous} propose to mine the modality interactions in test-time training.
Recently, Zhang \emph{et al.}.~\cite{zhang2024magic} propose to select diverse tokens and suppress the influence of backgrounds.
Although these methods achieve remarkable performance, they often overlook the dynamic quality changes in multi-modal imaging.
Meanwhile, they lack the ability to adaptively balance the multi-modal features based on instance characteristics.
In contrast, our porposed DeMo can effectively address these issues by adaptively weighting decoupled features, enhancing the model's robustness in complex scenarios.
\subsection{Mixtures of Experts}
The Mixture of Experts (MoE)~\cite{jacobs1991adaptive} is designed to tackle complex tasks by combining the specialized knowledge of multiple experts.
Recently, MoE has advanced significantly in various fields, including natural language processing~\cite{dai2024deepseekmoe}, computer vision~\cite{hwang2023tutel,chowdhury2023patch} and multi-modal learning~\cite{li2024uni,lin2024moe}.
In object ReID, MoE has been applied to domain generalizable ReID~\cite{xu2022mimic,kuang2024unity} and unsupervised ReID~\cite{li2023multi}, but its potential in multi-modal object ReID remains unexplored.
Meanwhile, many approaches~\cite{chen2024emoe,gui2024eegmamba} directly apply MoE without explicitly decoupling the expert inputs, which may lead to feature entanglement and limit the effectiveness of MoE.
In contrast, we perform hierarchical decoupling of multi-modal features, providing MoE with more flexible and specialized expert inputs.
Besides, existing methods~\cite{liu2024completed} often rely on simple techniques to generate gating weights, which may not fully capture the intricate relationships between experts.
To address this issue, we introduce ATMoE, which replaces traditional gating methods with attention-guided interactions between decoupled features.
It ensures more accurate and context-aware expert weighting, enhancing the model's adaptability to dynamic imaging conditions.
%
%
\begin{figure*}[tb]
    \centering
    \includegraphics[width=1\textwidth]{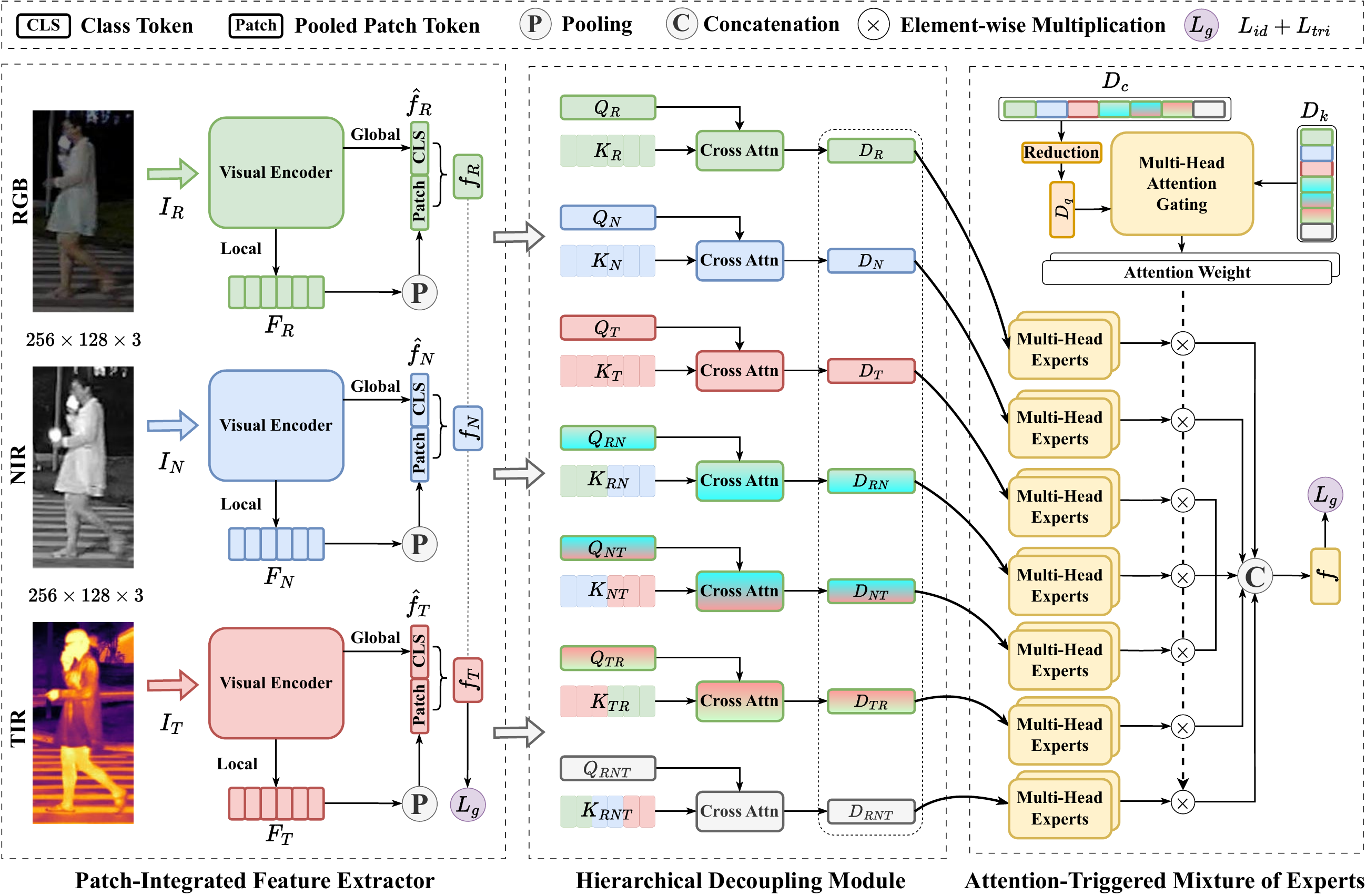}
    \caption{The overall framework of our DeMo.
    We first employ a Patch-Integrated Feature Extractor (PIFE) to extract multi-granularity features from each modality.
    Then, the Hierarchical Decoupling Module (HDM) decouples multi-modal features into different levels with learnable query tokens.
    Finally, the Attention-Triggered Mixture of Experts (ATMoE) adaptively balances the decoupled features with accurate and context-aware weights, generating robust multi-modal features.}
    \label{fig:Overall}
    \vspace{-0mm}
    \end{figure*}
\section{Methodology}
As shown in Fig.~\ref{fig:Overall}, our proposed DeMo is composed of three main components: Patch-Integrated Feature Extractor (PIFE), Hierarchical Decoupling Module (HDM) and Attention-Triggered Mixture of Experts (ATMoE).
%
\subsection{Patch-Integrated Feature Extractor}
To fully extract discriminative information from different modalities, we propose a Patch-Integrated Feature Extractor (PIFE) to capture multi-granularity features from each modality.
More specifically, the multi-modal inputs \(I_{m}\) (\(m \in \{R, N, T\}\)) are fed into the visual encoder \(\varTheta\) to produce patch tokens \(F_{m} \in \mathbb{R}^{N_{p} \times C}\) and class token \(\hat{f}_{m} \in \mathbb{R}^{C}\):
\begin{equation}
    F_{m}, \hat{f}_{m} = \varTheta(I_{m}).
\end{equation}
Here, \(R\), \(N\) and \(T\) represent the RGB, NIR and TIR modalities, respectively.
\(N_{p}\) denotes the number of patch tokens and \(C\) is the embedding dimension.
As shown in the left part of Fig.~\ref{fig:Overall}, we pool the patch tokens \(F_{m}\) and concatenate them with the corresponding class token \(\hat{f}_{m}\).
The concatenated features are then passed through a projection layer to obtain the modality-specific features \(f_{m}\) as follows:
\begin{equation}
    f_{m} = \omega ( W_\text{pro} ( \mathrm{LN} ( [\hat{f}_{m}, P(F_{m})] ))) ,
\end{equation}
where \(P(\cdot)\) is the average pooling operation and \([\cdot]\) means concatenation.
\(\mathrm{LN}(\cdot)\) represents the layer normalization~\cite{ba2016layer}.
\(W_\text{pro} \in \mathbb{R}^{2C \times C}\) is the projection matrix.
\(\omega(\cdot)\) is the GELU activation function~\cite{hendrycks2016gaussian}.
By integrating global and local information, we obtain multi-granularity features for each modality, enhancing the subsequent multi-modal fusion.
\subsection{Hierarchical Decoupling Module}
Multi-modal features consist of both modality-specific and modality-shared information.
Previous methods~\cite{wang2023top,wang2024heterogeneous,zhang2024magic} often neglect the mutual interference between modalities, leading to weakened modality-specific features and decreased diversity.
To address these issues, we propose a Hierarchical Decoupling Module (HDM).
As shown in Fig.~\ref{fig:Overall}, the decoupling process involves comprehensive interactions using cross-attentions.
Specifically, the HDM can be divided into three processes: unimodal-specific, bimodal-shared and trimodal-shared feature decoupling.
Details of each process are as follows.
\\
\textbf{Unimodal-specific Feature Decoupling.}
In the first three rows of HDM in Fig.~\ref{fig:Overall}, we show the unimodal-specific feature decoupling process.
For each modality, we first initialize a learnable query token \(Q_{m_{1}} \in \mathbb{R}^{C}\) and key tokens \(K_{m_{1}} \in \mathbb{R}^{(N_{p}+1) \times C}\), where \(m_{1} \in \{R, N, T\}\).
Here, key tokens \(K_{m_{1}}\) are constructed by concatenating the enhanced token and patch tokens from the corresponding modality $m_{1}$:
\begin{equation}
    K_{m_{1}} = [{f}_{m_{1}}, F_{m_{1}}].
\end{equation}
Then, \(Q_{m_{1}}\) is used to interact with \(K_{m_{1}}\) to obtain the decoupled unimodal-specific feature \(D_{m_{1}}\) as follows:
\begin{equation}
    D_{m_{1}} = \varPhi(Q_{m_{1}}, K_{m_{1}}),
\end{equation}
where $\varPhi$ denotes the multi-head cross-attention mechanism~\cite{vaswani2017attention}.
This approach leverages the learnable query token \(Q_{m_{1}}\) to dynamically focus on and highlight crucial modality-specific information, enabling a refined and context-aware extraction of unimodal features.
\\
\textbf{Bimodal-shared Feature Decoupling.}
As shown in the 4-6 rows of HDM in Fig.~\ref{fig:Overall}, we illustrate the bimodal-shared feature decoupling process.
Similar to the unimodal-specific feature decoupling, we generate a learnable query token \(Q_{m_{2}} \in \mathbb{R}^{C}\) and key tokens \(K_{m_{2}} \in \mathbb{R}^{(2N_{p}+2) \times C}\) for paired modalities \(m_{2} \in \{RN, NT, TR\}\).
Here, the key tokens \(K_{m_{2}}\) are constructed by concatenating all tokens from the corresponding paired modalities \(m_{2}\) as follows:
\begin{equation}
    K_{m_{2}} = [{f}_{m_{2}[0]}, F_{m_{2}[0]}, {f}_{m_{2}[1]}, F_{m_{2}[1]}],
\end{equation}
where \(m_{2}[0]\) and \(m_{2}[1]\) represent the two modalities in the paired modalities \(m_{2}\).
If \(m_{2} = RN\), then \(m_{2}[0] = R\) and \(m_{2}[1] = N\).
After that, we use \(Q_{m_{2}}\) to extract the decoupled bimodal-shared feature \(D_{m_{2}}\) from \(K_{m_{2}}\) as follows:
\begin{equation}
    D_{m_{2}} = \varPhi(Q_{m_{2}}, K_{m_{2}}).
\end{equation}
Through this interaction, the learnable query token \(Q_{m_{2}}\) integrates discriminative information from paired modalities, enhancing the representation of modality-shared features.
\\
\textbf{Trimodal-shared Feature Decoupling.}
In the last row of HDM in Fig.~\ref{fig:Overall}, we show the trimodal-shared feature decoupling process.
The only difference is that the key token \(K_{m_{3}} \in \mathbb{R}^{(3N_{p}+3) \times C}\), where \(m_{3} = RNT\), is constructed by concatenating all tokens from three modalities as follows:
\begin{equation}
    \scalebox{0.99}{$K_{m_{3}} = [{f}_{m_{3}[0]}, F_{m_{3}[0]}, {f}_{m_{3}[1]}, F_{m_{3}[1]}, {f}_{m_{3}[2]}, F_{m_{3}[2]}]$},
\end{equation}
where \(m_{3}[0]\), \(m_{3}[1]\) and \(m_{3}[2]\) represent the $R$, $N$ and $T$ modalities, respectively.
Then, we use \(Q_{m_{3}}\) to extract the decoupled trimodal-shared feature \(D_{m_{3}}\) from \(K_{m_{3}}\) as:
\begin{equation}
    D_{m_{3}} = \varPhi(Q_{m_{3}}, K_{m_{3}}).
\end{equation}
If there are highly shared discriminative regions among three modalities, the query token will prioritize these regions, assigning higher weights to them.
%
%
Thus, cross-attention helps \(D_{m_{3}}\) better capture shared discriminative information.

Finally, we obtain the decoupled features \(D_{m_{1}}\), \(D_{m_{2}}\) and \(D_{m_{3}}\) for unimodal-specific, bimodal-shared and trimodal-shared information, respectively.
By separating modality-specific and modality-shared information, the model can prevent interferences between modalities, preserving each modality's unique strengths and enhancing feature diversity.
Additionally, it provides the MoE with more options to select the most suitable experts under varying imaging conditions, thereby improving the model's generalization ability.
\begin{figure}[tb]
    \centering
    \includegraphics[width=0.46\textwidth]{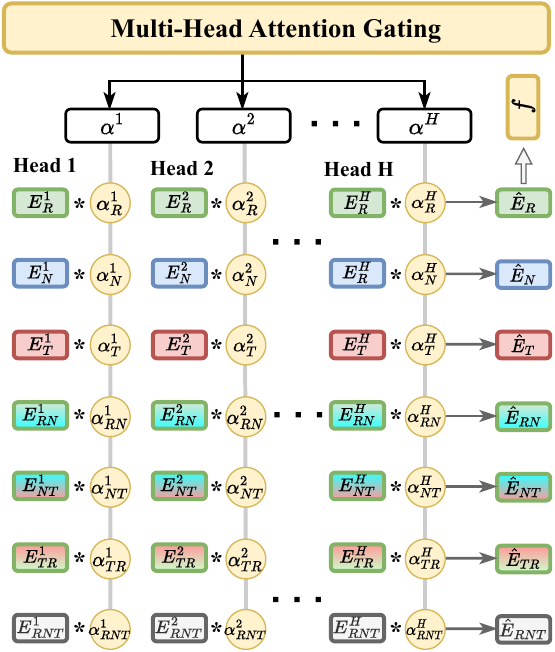}
    \vspace{-0mm}
    \caption{Detailed structure of ATMoE.}
    \label{fig:ATMoE}
\end{figure}
\subsection{Attention-Triggered Mixture of Experts}
To address the dynamic imaging quality and appropriately balance decoupled features across different instances, we introduce an Attention-Triggered Mixture of Experts (ATMoE).
Unlike traditional MoE~\cite{liu2024completed} where weights are directly generated from decoupled features, we incorporate an attention mechanism.
With attention-guided interactions between the integrated feature and each decoupled feature, ATMoE can assign more accurate and context-aware weights to the decoupled experts.
To be specific, as depicted in the top part of ATMoE in Fig.~\ref{fig:Overall}, we first concatenate the decoupled features and send them to a reduction layer $\varphi$ to obtain the query $D_{q} \in \mathbb{R}^{C}$ as follows:
\begin{equation}
    D_{q} = \varphi([D_{R}, D_{N}, D_{T},D_{RN}, D_{NT}, D_{TR}, D_{RNT}]),
\end{equation}
\begin{equation}
    \varphi(\mathcal{X}) = \mathrm{BN}(\omega(W_\text{red}(\mathcal{X}))),
\end{equation}
where $W_\text{red} \in \mathbb{R}^{n_{d}C \times C}$ is the reduction matrix and $\mathrm{BN}(\cdot)$ denotes the batch normalization~\cite{ioffe2015batch}.
Here, $n_{d}$ represents the number of decoupled features.
Meanwhile, we stack the decoupled features to generate the key tokens $D_{k} \in \mathbb{R}^{n_{d} \times C}$.
Then, we project $D_{q}$ and $D_{k}$ to generate $Q \in \mathbb{R}^{C}$ and $K \in \mathbb{R}^{n_{d} \times C}$ as follows:
\begin{equation}
    Q = W_{q}D_{q}, K = W_{k}D_{k},
\end{equation}
where $W_{q} \in \mathbb{R}^{C \times C}$ and $W_{k} \in \mathbb{R}^{C \times C}$ are the projection matrices.
After that, we use the multi-head attention mechanism to generate weights {\(A \in \mathbb{R}^{H \times n_{d}}\)} as follows:
\begin{equation}
    A = \left[\alpha^{1}, \alpha^{2}, \cdots, \alpha^{H}\right],
    \end{equation}
where $H$ is the number of heads and $\alpha^{h} \in \mathbb{R}^{n_{d}}$ is the attention weight for the $h$-th head.
To be specific, the attention weight $\alpha^{h}$ is calculated as follows:
\begin{equation}
    \alpha^{h} = \delta \left(\frac{Q^{h}K^{h\top}}{\sqrt{c}}\right).
\end{equation}
Here, $\delta(\cdot)$ is the softmax function and $c = \frac{C}{H}$ is the dimension of the head.
${{Q}^h} \in \mathbb{R}^{c}$ and ${{K}^h} \in \mathbb{R}^{n_{d} \times c}$ are the query and key tokens for the $h$-th head, respectively.
%

As shown in Fig.~\ref{fig:ATMoE}, after sending each decoupled feature to the corresponding expert, we chunk the experts' output into $H$ parts and multiply them with the corresponding attention weights.
Without loss of generality, we take $D_{R}$ as an example to obtain the expert output $E_{R}$ as follows:
\begin{equation}
    E_{R} = \mathrm{BN}(\omega(W_\text{exp}(D_{R}))),
\end{equation}
where $W_\text{exp} \in \mathbb{R}^{C \times C}$ is the expert matrix.
Then, we chunk $E_{R}$ into $H$ parts and multiply them with the corresponding attention weights to generate the weighted experts $\hat{E}_{R}$:
\begin{equation}
    \hat{E}_{R} = [E_{R}^{1}*\alpha^{1}_{R}, E_{R}^{2}*\alpha^{2}_{R}, \cdots, E_{R}^{H}*\alpha^{H}_{R}],
\end{equation}
where $E_{R}^{h} \in \mathbb{R}^{N_{p} \times c}$ is the $h$-th chunk of $E_{R}$.
Similarly, we can obtain weighted experts for other decoupled features.
Finally, we concatenate the outputs of all the weighted experts to form the final feature \( f \in \mathbb{R}^{n_{d}C} \).
Through the attention-guided interactions and the multi-head machanism, ATMoE can assign more accurate and context-aware weights to the decoupled experts, enhancing the feature robustness.
\subsection{Objective Functions}
As shown in Fig.~\ref{fig:Overall}, we optimize the model using multiple losses.
Features after PIFE and ATMoE are supervised by the label smoothing cross-entropy loss~\cite{szegedy2016rethinking} and triplet loss~\cite{hermans2017defense} as:
\begin{equation}
    \mathcal{L}_{g}(\mathcal{X}) = \mathcal{L}_{cross}(\mathcal{X}) + \mathcal{L}_{triplet}(\mathcal{X}),
\end{equation}
where $\mathcal{X}$ represents input features for supervision.
Finally, the overall loss $\mathcal{L}$ for our framework can be given by:
\begin{equation}
\mathcal{L} = \mathcal{L}_{g}([f_{R},f_{N},f_{T}]) + \mathcal{L}_{g}(f).
\end{equation}
\section{Experiments}
\subsection{Datasets and Evaluation Protocols}
\textbf{Datasets.}
We evaluate the proposed method on three multi-modal object ReID benchmarks.
To be specific, RGBNT201~\cite{zheng2021robust} is a multi-modal person ReID dataset, consisting of 4,787 aligned RGB, NIR and TIR images from 201 identities.
RGBNT100~\cite{li2020multi} is a large-scale multi-modal vehicle ReID dataset with 17,250 image triples, covering a wide range of challenging visual conditions.
MSVR310~\cite{zheng2022multi} is a small-scale multi-modal vehicle ReID dataset with 2,087 image triples, featuring high-quality images captured across diverse environments and time spans.
\\
\textbf{Evaluation Protocols.}
We use the mean Average Precision (mAP) and Cumulative Matching Characteristics (CMC) at Rank-K ($K=1, 5, 10$) to assess performance and present trainable parameters and FLOPs for complexity analysis.
\begin{table}[t]
    \centering
    \renewcommand\arraystretch{1.12}
    \setlength\tabcolsep{4pt}
    \resizebox{0.436\textwidth}{!}
    {
    \begin{tabular}{cccccc}
        \noalign{\hrule height 1pt}
    &\multicolumn{1}{c}{\multirow{1}{*}{\textbf{Methods}}}    & \textbf{mAP} & \textbf{R-1} & \textbf{R-5} & \textbf{R-10} \\ \hline
    \multirow{4}{*}{\rotatebox{90}{\textbf{Single}}}
    &MUDeep~\cite{qian2017multi} & 23.8 & 19.7 & 33.1 & 44.3 \\
    &OSNet~\cite{zhou2019omni}  & 25.4 & 22.3 & 35.1 & 44.7 \\
    &CAL~\cite{rao2021counterfactual}  & 27.6 & 24.3 & 36.5 & 45.7 \\
    &PCB~\cite{sun2018beyond}  & 32.8 & 28.1 & 37.4 & 46.9 \\ \hline
    \multirow{12}{*}{\rotatebox{90}{\textbf{Multi}}}
    & HAMNet~\cite{li2020multi}   & 27.7         & 26.3            & 41.5            & 51.7             \\
    & PFNet~\cite{zheng2021robust}    & 38.5         & 38.9            & 52.0            & 58.4             \\
    & DENet~\cite{zheng2023dynamic}    & 42.4         & 42.2            & 55.3            & 64.5            \\
    & IEEE~\cite{wang2022interact}     & 47.5         & 44.4            & 57.1            & 63.6             \\
    & LRMM~\cite{wu2025lrmm} & 52.3 & 53.4 & 64.6 & 73.2\\
    & UniCat$^*$~\cite{crawford2023unicat}   & 57.0         & 55.7            & -            & -            \\
    & HTT$^*$~\cite{wang2024heterogeneous} &71.1 &73.4 &83.1 &87.3\\
    & TOP-ReID$^*$~\cite{wang2023top}  &72.3 &76.6 &84.7 &89.4\\
    & EDITOR$^*$~\cite{zhang2024magic} & 66.5       & 68.3           & 81.1        & 88.2             \\
    & RSCNet$^*$~\cite{yu2024representation} & 68.2 & 72.5 & - & - \\
    & $\mathrm{\textbf{DeMo}}^*$  &\underline{73.7} 	 &\underline{80.5} 	 &\underline{88.3} 	 &\underline{91.5}      \\
    \rowcolor[gray]{0.92}
    & $\mathrm{\textbf{DeMo}}^\dagger$  &\textbf{79.0} 	 &\textbf{82.3} 	 &\textbf{88.8} 	 &\textbf{92.0}      \\
    \noalign{\hrule height 1pt}
    \end{tabular}
    }
    \vspace{-2mm}
    \caption{Performance comparison on RGBNT201.
    The best and second results are in bold and underlined, respectively.
    The symbol $\dagger$ denotes CLIP-based methods, $*$ indicates ViT-based methods and others are CNN-based methods.}
    \label{tab:multi-spectral person ReID}
    \vspace{-4mm}
\end{table}
\begin{table}[t]
    \centering
    \renewcommand\arraystretch{1.2}
    \setlength\tabcolsep{5pt}
    \resizebox{0.436\textwidth}{!}
    {
    \begin{tabular}{cccccc}
        \noalign{\hrule height 1pt}
    &\multicolumn{1}{c}{\multirow{2}{*}{\textbf{Methods}}} &  \multicolumn{2}{c}{\textbf{RGBNT100}} & \multicolumn{2}{c}{\textbf{MSVR310}} \\\cmidrule(r){3-4} \cmidrule(r){5-6}
    & & \textbf{mAP} & \textbf{R-1} & \textbf{mAP} & \textbf{R-1} \\
    \hline
    \multirow{5}{*}{\rotatebox{90}{\textbf{Single}}}
    &PCB~\cite{sun2018beyond}& 57.2 & 83.5 & 23.2 & 42.9 \\
    &BoT~\cite{luo2019bag} & 78.0 & 95.1 & 23.5 & 38.4 \\
    &OSNet~\cite{zhou2019omni}& 75.0 & 95.6 & 28.7 & 44.8 \\
    &AGW~\cite{ye2021deep} & 73.1 & 92.7 & 28.9 & 46.9 \\
    &TransReID$^*$~\cite{he2021transreid}& 75.6 & 92.9 & 18.4 & 29.6 \\
    \hline
    \multirow{13}{*}{\rotatebox{90}{\textbf{Multi}}}
    &GAFNet~\cite{guo2022generative} & 74.4 & 93.4 & - & - \\
    &GPFNet~\cite{he2023graph} & 75.0 & 94.5 & - & - \\
    &PFNet~\cite{zheng2021robust}& 68.1 & 94.1 & 23.5 & 37.4 \\
    &HAMNet~\cite{li2020multi} & 74.5 & 93.3 & 27.1 & 42.3 \\
    &CCNet~\cite{zheng2022multi} & 77.2 & 96.3 & 36.4 & \underline{55.2} \\
    & LRMM~\cite{wu2025lrmm} & 78.6 & \underline{96.7} & 36.7 &49.7\\
    &PHT$^*$~\cite{pan2023progressively} & 79.9 & 92.7 & - & - \\
    & HTT$^*$~\cite{wang2024heterogeneous} &75.7&92.6&- &-\\
    & TOP-ReID$^*$~\cite{wang2023top} &81.2 & 96.4 & 35.9 & 44.6 \\
    & EDITOR$^*$~\cite{zhang2024magic} & 82.1 & 96.4 &39.0 & 49.3\\
    & RSCNet$^*$~\cite{yu2024representation} &82.3 &96.6 &\underline{39.5} &49.6\\
    & $\mathrm{\textbf{DeMo}}^*$&\underline{82.4} &96.0 & 39.1&48.6\\
    \rowcolor[gray]{0.92}
    & $\mathrm{\textbf{DeMo}}^\dagger$& \textbf{86.2} 	&\textbf{97.6} &\textbf{49.2}	&\textbf{59.8} \\
    \noalign{\hrule height 1pt}
    \end{tabular}
    }
    \vspace{-2mm}
    \caption{Performance on RGBNT100 and MSVR310.}
    \label{tab:multi-spectral vehicle ReID}
    \vspace{-4mm}
\end{table}
\subsection{Implementation Details}
Our model is implemented using PyTorch with an NVIDIA A100 GPU.
We use the pre-trained ViT~\cite{dosovitskiy2020image} or CLIP~\cite{radford2021learning} as the visual encoder.
The number of experts $n_{d}$ is set to 7.
Images in triples are resized to 256$\times$128 for RGBNT201 and 128$\times$256 for RGBNT100/MSVR310.
For data augmentation, we apply random horizontal flipping, cropping and erasing~\cite{zhong2020random}.
For RGBNT201 and MSVR310, the mini-batch size is set to 64, sampling 8 images per identity.
For RGBNT100, the mini-batch size is 128 with 16 images per identity.
We fine-tune the proposed modules using the Adam optimizer with a learning rate of 3.5$\mathrm{e}^{-4}$ and a smaller learning rate of 5$\mathrm{e}^{-6}$ for the visual encoder.
The total number of training epochs is 50.
The detailed configurations and results are available at https://github.com/924973292/DeMo.
\begin{table*}[t]
    \centering
    \renewcommand\arraystretch{1.0}
    \setlength\tabcolsep{5pt}
    \resizebox{1\textwidth}{!}
    {
    \begin{tabular}{ccccccccccccccc}
        \noalign{\hrule height 1pt}
    \multicolumn{1}{c}{\multirow{2}{*}{\textbf{Methods}}} &  \multicolumn{2}{c}{\textbf{M (RGB)}} & \multicolumn{2}{c}{\textbf{M (NIR)}} & \multicolumn{2}{c}{\textbf{M (TIR)}} & \multicolumn{2}{c}{\textbf{M (RGB+NIR)}} & \multicolumn{2}{c}{\textbf{M (RGB+TIR)}} & \multicolumn{2}{c}{\textbf{M (NIR+TIR)}} & \multicolumn{2}{c}{\textbf{Average}} \\
    \cmidrule(r){2-3} \cmidrule(r){4-5} \cmidrule(r){6-7} \cmidrule(r){8-9} \cmidrule(r){10-11} \cmidrule(r){12-13} \cmidrule(r){14-15}
        & \textbf{mAP} & \textbf{R-1}  & \textbf{mAP} & \textbf{R-1} & \textbf{mAP} & \textbf{R-1} & \textbf{mAP} & \textbf{R-1} & \textbf{mAP} & \textbf{R-1} & \textbf{mAP} & \textbf{R-1} & \textbf{mAP} & \textbf{R-1} \\
        \hline
    PCB  & 23.6 & 24.2 & 24.4 & 25.1 & 19.9 & 14.7 & 20.6 & 23.6 & 11.0 & 6.8 & 18.6 & 14.4 & 19.7 & 18.1 \\
    TOP-ReID & \underline{54.4} & \underline{57.5} & \underline{64.3} & \underline{67.6} & \underline{51.9} & \textbf{54.5} & \underline{35.3} &\underline{35.4} & \underline{26.2} & \textbf{26.0} & \underline{34.1} & \underline{31.7} & \underline{44.4} & \underline{45.4} \\
    \rowcolor[gray]{0.92}
    \textbf{DeMo} & \textbf{63.3} & \textbf{65.3} & \textbf{72.6} & \textbf{75.7} & \textbf{56.2} & \underline{54.1} & \textbf{45.6} & \textbf{46.5} & \textbf{26.3} & \underline{24.9} & \textbf{40.3} & \textbf{38.5} & \textbf{50.7} & \textbf{50.8} \\
        \noalign{\hrule height 1pt}
    \end{tabular}
    }
    \vspace{-2mm}
    \caption{Performance of missing-modality settings on RGBNT201. “M (X)" means missing the X image modality.}
    \label{tab:missing-modality person ReID}
    \vspace{-2mm}
    \end{table*}
    \begin{table*}[t]
        \centering
        \renewcommand\arraystretch{1.0}
        \setlength\tabcolsep{5pt}
        \resizebox{1\textwidth}{!}
        {
        \begin{tabular}{cccccccccccccccc}
            \noalign{\hrule height 1pt}
        \multicolumn{1}{c}{\multirow{2}{*}{\textbf{Methods}}} &  \multicolumn{2}{c}{\textbf{M (RGB)}} & \multicolumn{2}{c}{\textbf{M (NIR)}} & \multicolumn{2}{c}{\textbf{M (TIR)}} & \multicolumn{2}{c}{\textbf{M (RGB+NIR)}} & \multicolumn{2}{c}{\textbf{M (RGB+TIR)}} & \multicolumn{2}{c}{\textbf{M (NIR+TIR)}} & \multicolumn{2}{c}{\textbf{Average}} \\
            \cmidrule(r){2-3} \cmidrule(r){4-5} \cmidrule(r){6-7} \cmidrule(r){8-9} \cmidrule(r){10-11} \cmidrule(r){12-13} \cmidrule(r){14-15}
        \cmidrule(r){2-3} \cmidrule(r){4-5} \cmidrule(r){6-7} \cmidrule(r){8-9} \cmidrule(r){10-11} \cmidrule(r){12-13}
        &\textbf{mAP} & \textbf{R-1}  & \textbf{mAP} & \textbf{R-1} & \textbf{mAP} & \textbf{R-1} & \textbf{mAP} & \textbf{R-1} & \textbf{mAP} & \textbf{R-1} & \textbf{mAP} & \textbf{R-1} & \textbf{mAP} & \textbf{R-1}\\
            \hline
            TOP-ReID & \underline{70.6} & \underline{90.6} & \underline{77.9} & \underline{94.5} & \underline{64.0} & \underline{81.5} & \underline{42.5} & \underline{69.3} & \underline{45.9} & \underline{65.4} & \underline{55.4} & \underline{77.8} & \underline{59.4} & \underline{79.9} \\

        \rowcolor[gray]{0.92}
        \textbf{DeMo} & \textbf{81.0}    & \textbf{94.5}    & \textbf{84.1}     & \textbf{96.5}     & \textbf{71.1}     & \textbf{87.6}     &\textbf{50.2}         & \textbf{73.7}      & \textbf{59.6}     & \textbf{78.1}     & \textbf{66.3}    & \textbf{82.8}    & \textbf{68.7} & \textbf{85.5} \\
            \noalign{\hrule height 1pt}
        \end{tabular}
        }
        \vspace{-2mm}
        \caption{Performance of missing-modality settings on RGBNT100.}
        \label{tab:missing-modality vehicle ReID}
        \vspace{-2mm}
    \end{table*}
\subsection{Comparison with State-of-the-Art Methods}
\textbf{Multi-modal Person ReID.}
In Tab.~\ref{tab:multi-spectral person ReID}, we compare our proposed DeMo with single-modal and multi-modal methods on RGBNT201.
Generally, multi-modal methods significantly outperform single-modal methods by integrating complementary information from different modalities.
Among them, models based on ViT and CLIP perform better than those based on CNNs.
Specifically, DeMo$^*$ shows a 1.4\% improvement in mAP and a 3.9\% improvement in Rank-1 compared with TOP-ReID$^*$.
This highlights DeMo's effectiveness in dynamically integrating multi-modal information in complex environments.
Additionally, DeMo$^\dagger$ utilizes CLIP's pre-trained knowledge to improve the robustness of multi-modal features, achieving a 6.7\% improvement in mAP and a 5.7\% improvement in Rank-1 over TOP-ReID$^*$.
These results confirm DeMo's capability in managing dynamic changes in modality fusion.
\\
\textbf{Multi-modal Vehicle ReID.}
As shown in Tab.~\ref{tab:multi-spectral vehicle ReID}, TransReID$^*$ achieves an mAP of 75.6\% on the large-scale RGBNT100 dataset.
However, on the smaller MSVR310 dataset, it performs worse than AGW and OSNet, which are better suited for small-scale datasets.
Among multi-modal methods, EDITOR$^*$ demonstrates significant improvements on both datasets.
Our DeMo$^*$, with its simpler structure, delivers competitive results and avoids the instability issues present in EDITOR$^*$.
Additionally, DeMo$^\dagger$ achieves 86.2\% mAP on RGBNT100, outperforming EDITOR$^*$ by 4.1\%.
On MSVR310, DeMo$^\dagger$ exceeds EDITOR$^*$ and TOP-ReID$^*$ by over 10.2\% in mAP and 10.5\% in Rank-1.
These results highlight DeMo's robustness in integrating multi-modal information across dynamic environments.
\\
\textbf{Multi-modal Object ReID with Missing Modalities.}
To assess DeMo's robustness in missing-modality scenarios, we conduct experiments on RGBNT201 and RGBNT100.
As shown in Tab.~\ref{tab:missing-modality person ReID} and Tab.~\ref{tab:missing-modality vehicle ReID}, DeMo consistently outperforms TOP-ReID in these settings.
Despite lacking specific designs like the reconstruction modules in TOP-ReID, DeMo achieves competitive performance through automatic feature weighting.
In all missing-modality settings, DeMo achieves an average mAP of 50.7\% on RGBNT201, which is 6.3\% higher than TOP-ReID.
On RGBNT100, DeMo achieves an average mAP of 68.7\%, surpassing TOP-ReID by 9.3\%.
These results fully validate the robustness of our proposed DeMo in handling diverse and complex ReID scenarios.
\begin{table}[t]
    \centering
    \renewcommand\arraystretch{1.0}
    \setlength\tabcolsep{4.5pt}
    \resizebox{0.47\textwidth}{!}
    {
    \begin{tabular}{cccccccc}
        \noalign{\hrule height 1pt}
        \multicolumn{1}{c}{\multirow{2}{*}{\textbf{Index}}}                   &\multicolumn{3}{c}{\textbf{Modules}}                                  & \multicolumn{2}{c}{\textbf{Metrics}} & \textbf{Params}                 & \textbf{FLOPs}                  \\
        \cmidrule(r){2-4} \cmidrule(r){5-6} \cmidrule(r){7-7} \cmidrule(r){8-8}
    & \textbf{PIFE}              & \textbf{HDM}                & \textbf{ATMoE}                   & \textbf{mAP}    & \textbf{R-1}   & \textbf{M}                      & \textbf{G}                      \\ \hline
    A                  & \ding{53}                  & \ding{53}                  & \ding{53}                    & 70.7  & 72.4 & 86.41                  & 34.28                  \\
    B                  & \ding{51}                  & \ding{53}                  & \ding{53}                      & 73.0  & 75.8 & 87.99                  & 34.28                  \\
    \multirow{1}{*}{C} & \multirow{1}{*}{\ding{51}} & \multirow{1}{*}{\ding{51}} & \multirow{1}{*}{\ding{53}}    & 74.4  & 77.5 & 95.96                  & 35.09 \\
    \multirow{1}{*}{D} & \multirow{1}{*}{\ding{51}} & \multirow{1}{*}{\ding{51}} & \multirow{1}{*}{\ding{51}}    & 76.8  & 79.8 & 98.79                & 35.10 \\
    \rowcolor[gray]{0.92}
    \multirow{1}{*}{E} & \multirow{1}{*}{\ding{51}} & \multirow{1}{*}{\ding{51}} & \multirow{1}{*}{\ding{51}}   &\textbf{79.0} 	 &\textbf{82.3}  & 98.79                 & 35.10 \\
    \noalign{\hrule height 1pt}
    \end{tabular}
    }
    \vspace{-2mm}
    \caption{Comparison with different modules.
    Model D infers with \( f \), while Model E further incorporates \( f_{m} \).}
    \label{tab:ablation}
\end{table}
\subsection{Ablation Studies}
We evaluate the effectiveness of different modules on the RGBNT201 dataset.
To be specific, our baseline model only utilizes the class tokens from the visual encoders.
\\
\textbf{Effects of Key Modules.}
Tab.~\ref{tab:ablation} shows the performance comparison with different modules.
Model A is the baseline model, achieving an mAP of 70.7\% and Rank-1 of 72.4\%.
With PIFE, Model B increases the performance to an mAP of 73.0\%.
Model C further incorporates HDM, boosting mAP to 74.4\% and Rank-1 to 77.5\%, indicating the robustness of decoupled multi-modal features.
Model D introduces ATMoE, delivering an mAP of 76.8\% and Rank-1 of 79.8\%.
Finally, Model E combines features after PIFE and ATMoE, achieving the best results with an mAP of 79.0\% and Rank-1 of 82.3\%.
As for the complexity analysis, our proposed modules introduce a minor increase in learnable parameters (less than 13MB).
In addition, the increase of FLOPs is rather small when compared with the baseline model.
These results demonstrate the effectiveness of our methods.
\\
\textbf{Effects of Gating Methods.}
Tab.~\ref{tab:sepvsunion} compares the performance of different gating methods.
The ``Simple'' method generates weights through a linear transformation and a softmax function applied to the decoupled features.
The direct addition of weighted experts in ``Simple$^{A}$'' leads to worse performances, while the simple concatenation in ``Simple$^{C}$'' achieves better results.
In contrast, our ATMoE leverages attention mechanisms to generate more accurate weights.
Especially, ATMoE with a single head achieves an mAP of 77.6\%, outperforming the ``Simple'' method.
Further improvements are observed with the number of heads increased to 2 and 4, resulting in mAP scores of 78.1\% and 79.0\%, respectively.
These results clearly highlight the effectiveness of ATMoE in generating more accurate weights.
\begin{table}[t]
    \centering
    \renewcommand\arraystretch{1.0}
    \setlength\tabcolsep{3.8pt}
    \resizebox{0.48\textwidth}{!}
    {
    \begin{tabular}{ccccccc}
        \noalign{\hrule height 1pt}
    \multicolumn{1}{c}{\multirow{1}{*}{\textbf{Gating}}}    &\textbf{Head} & \textbf{mAP} & \textbf{Rank-1} & \textbf{Rank-5} & \textbf{Rank-10} \\
    \hline
    Simple$^{A}$&-&  75.2         & 76.7            & 84.6           & 89.7             \\
    Simple$^{C}$&-&  76.7         & 77.4            & 85.2           & 90.1             \\
    Attention$^{C}$&1  & 77.6	& 81.6	& 87.7	& 90.3   \\
    Attention$^{C}$&2&78.1 	 &81.8 	 &\underline{88.4} 	 &\underline{91.7}             \\
    \rowcolor[gray]{0.92}
    Attention$^{C}$&4&\textbf{79.0} 	 &\underline{82.3} 	 &\textbf{88.8} 	 &\textbf{92.0}             \\
    Attention$^{C}$&8& \underline{78.2} & \textbf{82.5} & 88.2 & 90.4 \\
    \noalign{\hrule height 1pt}
    \end{tabular}
    }
    \vspace{-2mm}
    \caption{Comparison of gating methods.
    The symbol $A$ and $C$ indicate addition and concatenation of weighted experts.}
    \label{tab:sepvsunion}
    \vspace{-3mm}
\end{table}
\begin{figure*}[t]
    \centering
    \includegraphics[width=0.94\textwidth]{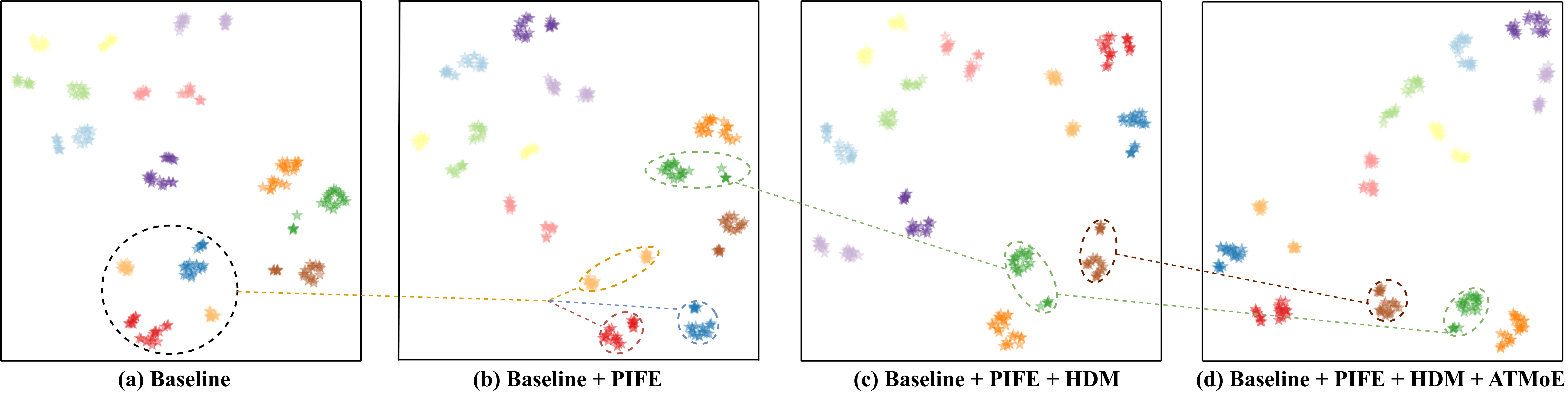}
    \vspace{-2mm}
    \caption{Feature distributions with t-SNE~\cite{van2008visualizing}.
    Different colors refer to different IDs.}
    \label{fig:tsne}
    \vspace{-2mm}
    \end{figure*}
    \begin{figure}[t]
    \centering
    \includegraphics[width=0.41\textwidth]{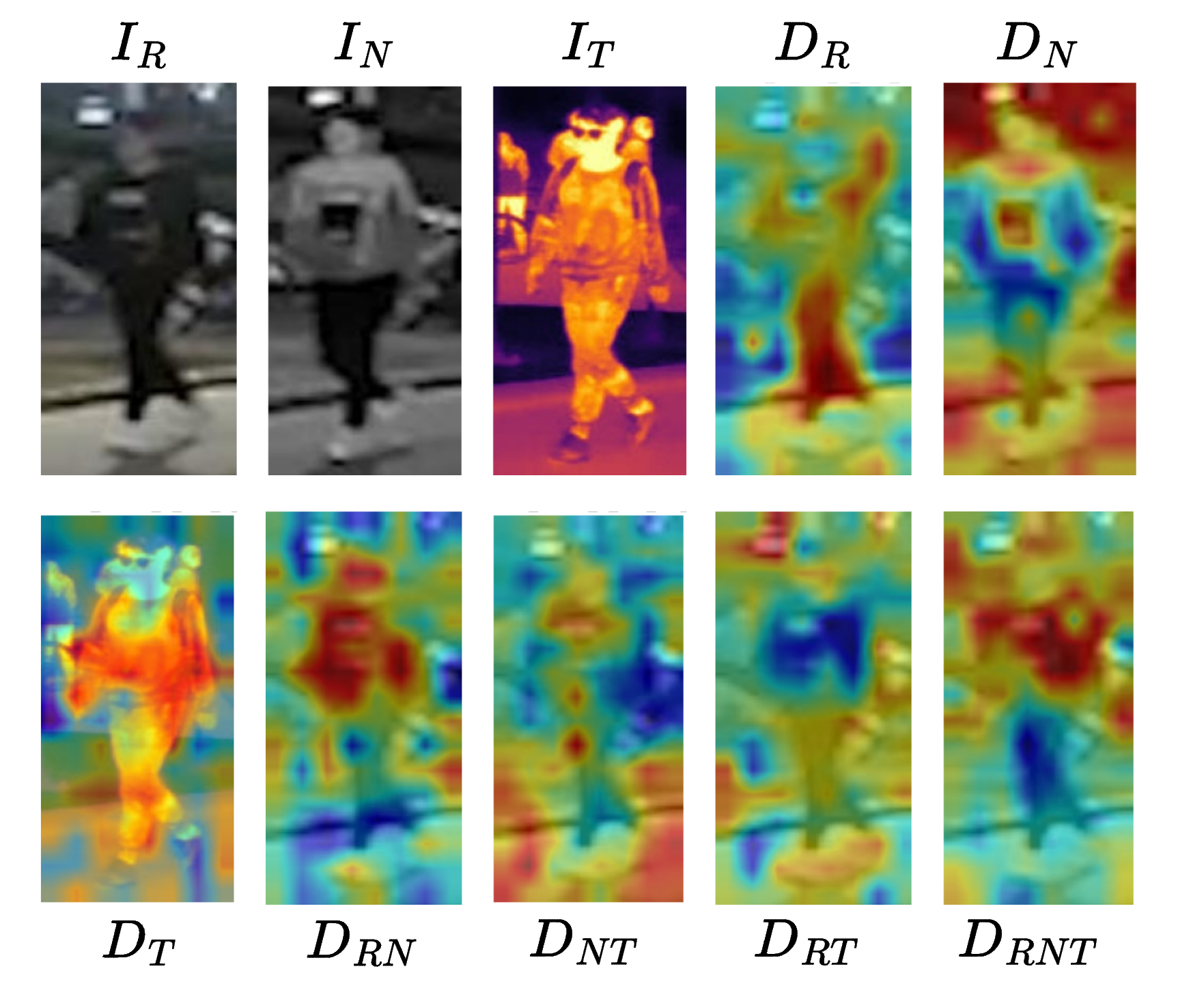}
    \vspace{-2mm}
    \caption{Activation maps of decoupled features.}
    \label{fig:activate}
    \vspace{-2mm}
    \end{figure}
    \begin{figure}[t]
    \centering
    \includegraphics[width=0.47\textwidth]{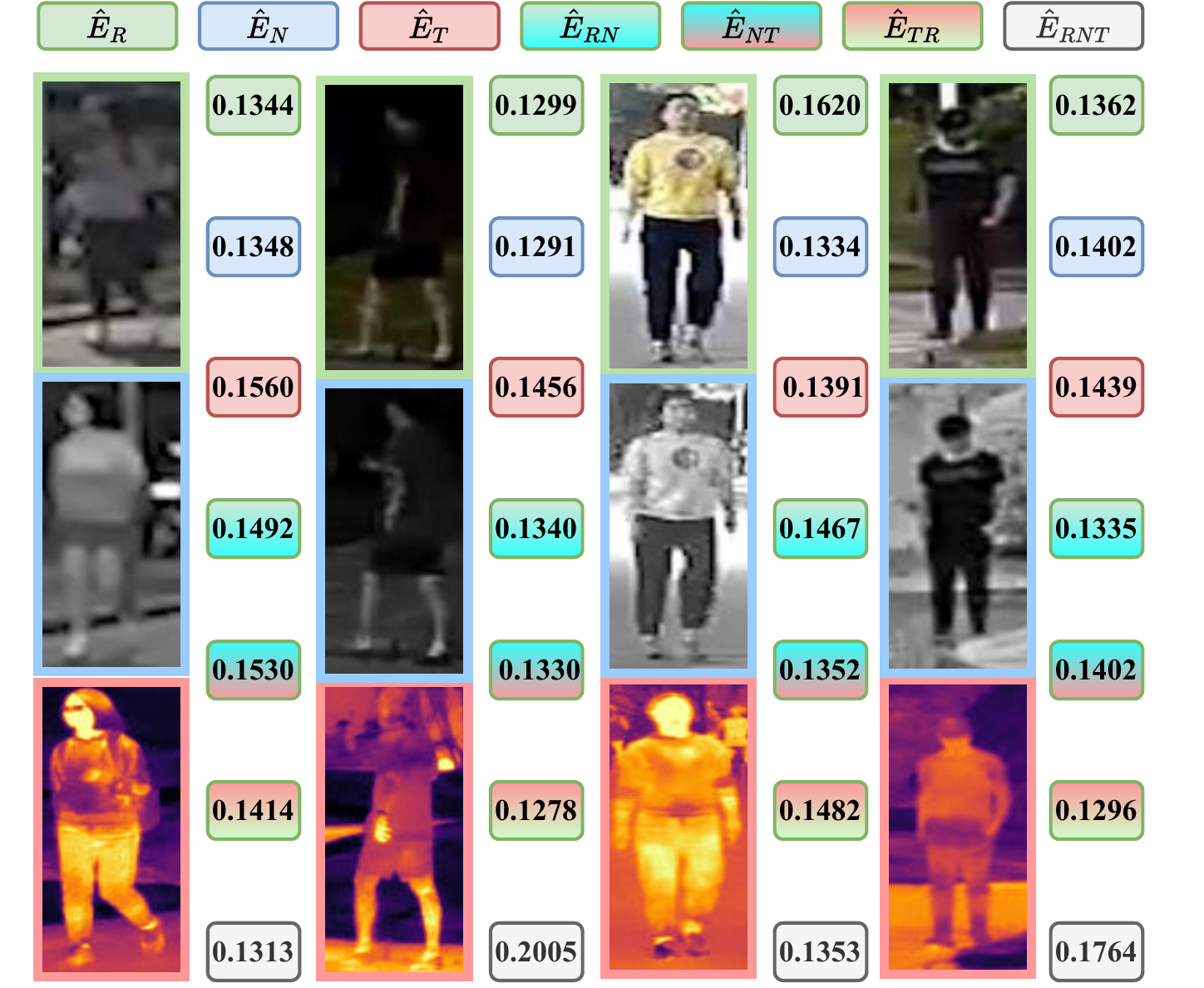}
    \caption{Visualization of dynamic weights across instances.
    Different colors correspond to distinct decoupled features.}
    \label{fig:dynamic}
    \vspace{-4mm}
    \end{figure}
\subsection{Visualization Analysis}
\textbf{Multi-modal Feature Distributions.}
In Fig.~\ref{fig:tsne}, we visualize the discriminative feature distributions of different modules.
From Fig.~\ref{fig:tsne} (a) to Fig.~\ref{fig:tsne} (b), challenging samples are better separated.
Compared with Fig.~\ref{fig:tsne} (b), HDM further narrows the distance between instances of the same ID, thereby enhancing feature discrimination, as shown in Fig.~\ref{fig:tsne} (c).
Finally, with ATMoE in Fig.~\ref{fig:tsne} (d), features for each ID become more compact and the gap between different IDs increases.
These visualizations fully validate the effectiveness of our proposed modules in improving feature discrimination.
\\
\textbf{Activation Maps of Decoupled Features.}
In Fig.~\ref{fig:activate}, we visualize the activation maps of decoupled features.
Different features focus on distinct regions of the input image.
Notably, $D_{RN}$ highlights areas that differ from those in $D_{R}$ and $D_{N}$, which are shared between $I_{R}$ and $I_{N}$.
This suggests that HDM effectively promotes the decoupling of multi-modal features.
Similar phenomena can be observed in the $D_{R}$, $D_{T}$ and $D_{RT}$ triplet, demonstrating HDM's effectiveness in enhancing multi-modal feature diversity.
\\
\textbf{Dynamic Weight Visualizations.}
As shown in Fig.~\ref{fig:dynamic}, we present the dynamic weights for different instances.
The weights of various decoupled features fluctuate across instances, highlighting the capability of ATMoE to adjust feature importance based on instance characteristics.
Notably, modalities that contain more details receive greater attention, enhancing the robustness against imaging variations.
\section{Conclusion}
In this paper, we present a novel framework named DeMo for multi-modal object ReID.
Our approach starts with a Patch-Integrated Feature Extractor (PIFE) to capture multi-granular features from diverse modalities.
Then, we introduce the Hierarchical Decoupling Module (HDM) to separate modality-specific information.
Finally, the Attention-Triggered Mixture of Experts (ATMoE) assigns accurate and context-aware weights to the decoupled experts.
DeMo effectively enhances feature robustness against variations in imaging quality across modalities.
Extensive experiments on three benchmarks validate the effectiveness of our DeMo.
\section{Acknowledgments}
This work was supported in part by the National Natural Science Foundation of China (No.62101092, 62476044, 62388101), Open Project of Anhui Provincial Key Laboratory of Multimodal Cognitive Computation, Anhui University (No.MMC202102, MMC202407) and Fundamental Research Funds for the Central Universities (No.DUT23BK050, DUT23YG232).
\bibliography{aaai25}
\section{A. Introduction}
In this supplementary material, we provide additional experimental details and visualizations to complement the main paper.
Specifically, the structure is organized as follows:
\begin{enumerate}
    \item \textbf{Module validation and hyper parameter analysis:}
    \begin{itemize}
        \item Analysis of different structures and hyper-parameters
        \item Model parameter comparison with other methods
        \item Generalization evaluation on vehicle datasets
    \end{itemize}

    \item \textbf{Extended visualizations for person and vehicle ReID:}
    \begin{itemize}
        \item Visualizations of dynamic weights in ATMoE
        \item Activation maps of decoupled features in HDM
        \item Rank list comparison across different modules
    \end{itemize}
\end{enumerate}
These analyses provide a comprehensive understanding of our DeMo and further validate the effectiveness of modules.
\begin{table}[t]
    \centering
    \renewcommand\arraystretch{1.2}
    \setlength\tabcolsep{3.8pt}
    \resizebox{0.47\textwidth}{!}
    {
    \begin{tabular}{ccccccccc}
        \noalign{\hrule height 1pt}
        \multicolumn{1}{c}{\multirow{2}{*}{\textbf{Index}}}&\multicolumn{1}{c}{\multirow{2}{*}{\textbf{Pooling Methods}}} &\multicolumn{4}{c}{\multirow{1}{*}{\textbf{RGBNT201}}}  & \textbf{Params} & \textbf{FLOPs} \\
    \cmidrule(r){3-6} \cmidrule(r){7-7} \cmidrule(r){8-8}
    & & \textbf{mAP} & \textbf{R-1} & \textbf{R-5} & \textbf{R-10}& \textbf{M} & \textbf{G} \\
    \hline
    \rowcolor[gray]{0.92}
    \textbf{A} &\textbf{Average Pooling} & \underline{79.0} 	 &\underline{82.3} 	 &\underline{88.8} 	 &\underline{92.0}      & 98.79                 & 35.10        \\
    B &Max Pooling &76.6 	&79.8 	&88.4 	&91.3   & 98.79                 & 35.10  \\
    C &GeM Pooling &\textbf{79.1} 	 &\textbf{83.7} 	 &\textbf{90.0} 	 &\textbf{92.7}      & 98.79                 & 35.10       \\
    \noalign{\hrule height 1pt}
    \end{tabular}
    }
    \caption{Comparison of pooling methods in PIFE.}
    \label{tab:pooling}
  \end{table}
  \begin{table}[t]
      \vspace{-0mm}
      \centering
      \renewcommand\arraystretch{1.2}
      \setlength\tabcolsep{3.8pt}
      \resizebox{0.47\textwidth}{!}
      {
          \begin{tabular}{ccccccccc}
              \noalign{\hrule height 1pt}
              \multicolumn{1}{c}{\multirow{2}{*}{\textbf{Index}}} & \multicolumn{1}{c}{\multirow{2}{*}{\textbf{Methods}}} & \multicolumn{4}{c}{\multirow{1}{*}{\textbf{RGBNT201}}} & \textbf{Params} & \textbf{FLOPs} \\
              \cmidrule(r){3-6} \cmidrule(r){7-7} \cmidrule(r){8-8}
              & & \textbf{mAP} & \textbf{R-1} & \textbf{R-5} & \textbf{R-10} & \textbf{M} & \textbf{G} \\ \hline
              A & w/o PIFE     & 77.7         & 80.1  &87.4 &90.7     & 97.21      & 35.10 \\
              \rowcolor[gray]{0.92}
              B & \textbf{DeMo}         & \textbf{79.0}         & \textbf{82.3}    &\textbf{88.8} &\textbf{92.0}   & 98.79      & 35.10 \\ \hline
              \noalign{\hrule height 1pt}
          \end{tabular}
      }
      \caption{Comparison of w or w/o PIFE in DeMo.}
      \label{tab:comparison_pife}
  \end{table}
  \begin{table}[t]
    \centering
    \renewcommand\arraystretch{1.2}
    \setlength\tabcolsep{3.8pt}
    \resizebox{0.47\textwidth}{!}
    {
    \begin{tabular}{ccccccccc}
        \noalign{\hrule height 1pt}
        \multicolumn{1}{c}{\multirow{2}{*}{\textbf{Index}}}&\multicolumn{1}{c}{\multirow{2}{*}{\textbf{Interaction}}} &\multicolumn{4}{c}{\multirow{1}{*}{\textbf{RGBNT201}}}  & \textbf{Params} & \textbf{FLOPs} \\
    \cmidrule(r){3-6} \cmidrule(r){7-7} \cmidrule(r){8-8}
    & & \textbf{mAP} & \textbf{R-1} & \textbf{R-5} & \textbf{R-10}& \textbf{M} & \textbf{G} \\
    \hline
    A &w/o Interaction & 75.6 & \underline{80.9} & 87.1 & 91.0 & 94.58 & 34.28 \\
    B &Cross Attention w/o $f_{m}$ &74.4 	&79.9 	&86.0 	&89.2   & 98.79    &35.09  \\
    \rowcolor[gray]{0.92}
    \textbf{C} &\textbf{Cross Attention}&\textbf{79.0} 	 &\textbf{82.3} 	 &\textbf{88.8} 	 &\textbf{92.0}      & 98.79                 & 35.10       \\
    D &Transformer Block & \underline{77.8} & 80.6 & \underline{88.1} & \underline{91.5} & 113.51 & 39.59 \\
    \noalign{\hrule height 1pt}
    \end{tabular}
    }
    \caption{Comparison of interaction mechanisms in HDM.}
    \label{tab:interaction}
  \end{table}
  \begin{table}[t]
    \vspace{-0mm}
    \centering
    \renewcommand\arraystretch{1.2}
    \setlength\tabcolsep{3.8pt}
    \resizebox{0.47\textwidth}{!}
    {
        \begin{tabular}{cccccccc}
            \noalign{\hrule height 1pt}
            \multicolumn{1}{c}{\multirow{2}{*}{\textbf{Index}}}&\multicolumn{1}{c}{\multirow{2}{*}{\textbf{HDM Structures}}} &\multicolumn{4}{c}{\multirow{1}{*}{\textbf{RGBNT201}}}  & \textbf{Params} & \textbf{FLOPs} \\
            \cmidrule(r){3-6} \cmidrule(r){7-7} \cmidrule(r){8-8}
            & & \textbf{mAP} & \textbf{R-1} & \textbf{R-5} & \textbf{R-10}& \textbf{M} & \textbf{G} \\ \hline
            A     & $7K_{RNT}$         & 72.6                       & 75.6 & 83.2 &88.0& 98.79     & 35.71\\
            \rowcolor[gray]{0.92}
            B & \textbf{DeMo}         & \textbf{79.0}         & \textbf{82.3}    &\textbf{88.8} &\textbf{92.0}   & 98.79      & 35.10 \\ \hline
            \noalign{\hrule height 1pt}
        \end{tabular}
    }
    \caption{Comparison with $7K_{RNT}$ in HDM.}
    \label{tab:HDM_7K}
\end{table}
\section{B. Experimental Analysis}
\subsection{Different Structures and Hyper Parameters}
\textbf{Effect of Pooling Methods in PIFE.}
In Tab.~\ref{tab:pooling}, we present a comparison of various pooling methods applied in PIFE.
Model A utilizes Average Pooling, which delivers competitive performance while maintaining a low computational cost, as it does not introduce additional parameters.
Model B, which employs Max Pooling, shows suboptimal results compared to the other approaches.
Model C incorporates GeM Pooling~\cite{radenovic2018fine}, achieving the highest performance with a mAP of 79.1\%.
Nevertheless, to maintain simplicity, we select Average Pooling as our default pooling method in all experiments.
\\
\textbf{Effect of PIFE in DeMo.}
In Tab.~\ref{tab:comparison_pife}, we investigate the impact of PIFE on the overall performance of DeMo.
Model A represents our final model without PIFE, achieving an mAP of 77.7\% and Rank-1 of 80.1\%.
Model B introduces PIFE, which boosts the mAP to 79.0\% and Rank-1 to 82.3\%.
These results demonstrate the effectiveness of PIFE in enhancing feature representation and improving ReID performance.
Meanwhile, Model A still achieves competitive performance, verifying the robustness of our proposed DeMo even without the local feature enhancement.
\\
\textbf{Effect of Interaction Mechanisms in HDM.}
As shown in Tab.~\ref{tab:interaction}, we compare different interaction mechanisms within the HDM framework.
In Model A, decoupled features are directly derived from the modality-specific features \( f_{m}, m \in \{R, N, T\} \).
Here, a linear reduction is applied to ensure that the concatenated features share the same dimensionality.
For instance, \( D_{RNT} \in \mathbb{R}^{C} \) is obtained by applying a linear layer to the concatenation of \( [f_{R}, f_{N}, f_{T}] \).
Model B employs a Cross Attention mechanism but omits the modality-specific features \( f_{m} \) from the key tokens during the decoupling process, thus restricting the interaction to local features only.
Model C implements the full Cross Attention mechanism, which is adopted in our final model. This approach fully integrates global features into the interaction process.
Model D substitutes the Cross Attention mechanism with a standard Transformer Block, providing an alternative method for feature interaction.
In terms of experimental results, Model A exhibits unsatisfactory performance, highlighting the necessity of interaction mechanisms.
Model B underperforms relative to Model C, emphasizing the critical role of global features in effective interaction.
Although Model D demonstrates competitive performance, the Cross Attention mechanism proves more effective and the Transformer Block introduces significant computational overhead (FLOPs).
Therefore, we select the Cross Attention mechanism with the inclusion of \( f_{m}, m \in \{R, N, T\} \) in our proposed HDM, balancing high performance with lower complexity.
\\
\textbf{Effect of Different HDM Structures.}
In Tab.~\ref{tab:HDM_7K}, we compare different interaction structures within HDM.
With the combination of different modalities, each query can attend to different levels of shared features from various keys, enabling the model to focus on more diverse regions.
By replacing the modality combinations with seven sets of $K_{RNT}$ interactions, both the FLOPs increase and performance degrades, with queries focusing on similar regions, which limits the model’s ability to decouple modality-specific information.
Beseides, the retrieval performance of the features extracted from HDM dropped by nearly 10\% in mAP, highlighting the critical role of feature decoupling in HDM and its impact on the subsequent ATMoE stage.
\\
\textbf{Effect of Expert Structures in ATMoE.}
In Tab.~\ref{tab:expert}, we compare various expert structures within ATMoE.
Overall, complex structures tend to perform worse than simpler ones, likely due to overfitting.
Model A utilizes a Simple structure, delivering competitive performance.
Model B adopts a Bottleneck structure, but it underperforms relative to Model A.
Model C, which introduces additional parameters and FLOPs, yields the poorest performance among the models.
Consequently, we use the Simple structure for our experts.
\begin{table}[t]
    \centering
    \renewcommand\arraystretch{1.2}
    \setlength\tabcolsep{3.8pt}
    \resizebox{0.47\textwidth}{!}
    {
    \begin{tabular}{ccccccccc}
        \noalign{\hrule height 1pt}
        \multicolumn{1}{c}{\multirow{2}{*}{\textbf{Index}}}&\multicolumn{1}{c}{\multirow{2}{*}{\textbf{Expert Structures}}} &\multicolumn{4}{c}{\multirow{1}{*}{\textbf{RGBNT201}}}  & \textbf{Params} & \textbf{FLOPs} \\
    \cmidrule(r){3-6} \cmidrule(r){7-7} \cmidrule(r){8-8}
    & & \textbf{mAP} & \textbf{R-1} & \textbf{R-5} & \textbf{R-10}& \textbf{M} & \textbf{G} \\
    \hline
    \rowcolor[gray]{0.92}
    \textbf{A} &\textbf{Simple} & \textbf{79.0} 	 &\textbf{82.3} 	 &\textbf{88.8} 	 &\textbf{92.0}      & 98.79                 & 35.0983        \\
    B &Bottleneck  &\underline{77.6} 	&\underline{80.5} 	&\underline{88.3} 	&\underline{91.3}   & 98.56                 & 35.0980  \\
    C &FFN &75.9 	 &77.5 	 &85.2 	 &89.1      & 102.02               & 35.1015       \\
    \noalign{\hrule height 1pt}
    \end{tabular}
    }
    \caption{Comparison of expert structures in ATMoE.}
    \label{tab:expert}
  \end{table}
\begin{figure}[t]
  \centering
  \includegraphics[width=0.474\textwidth]{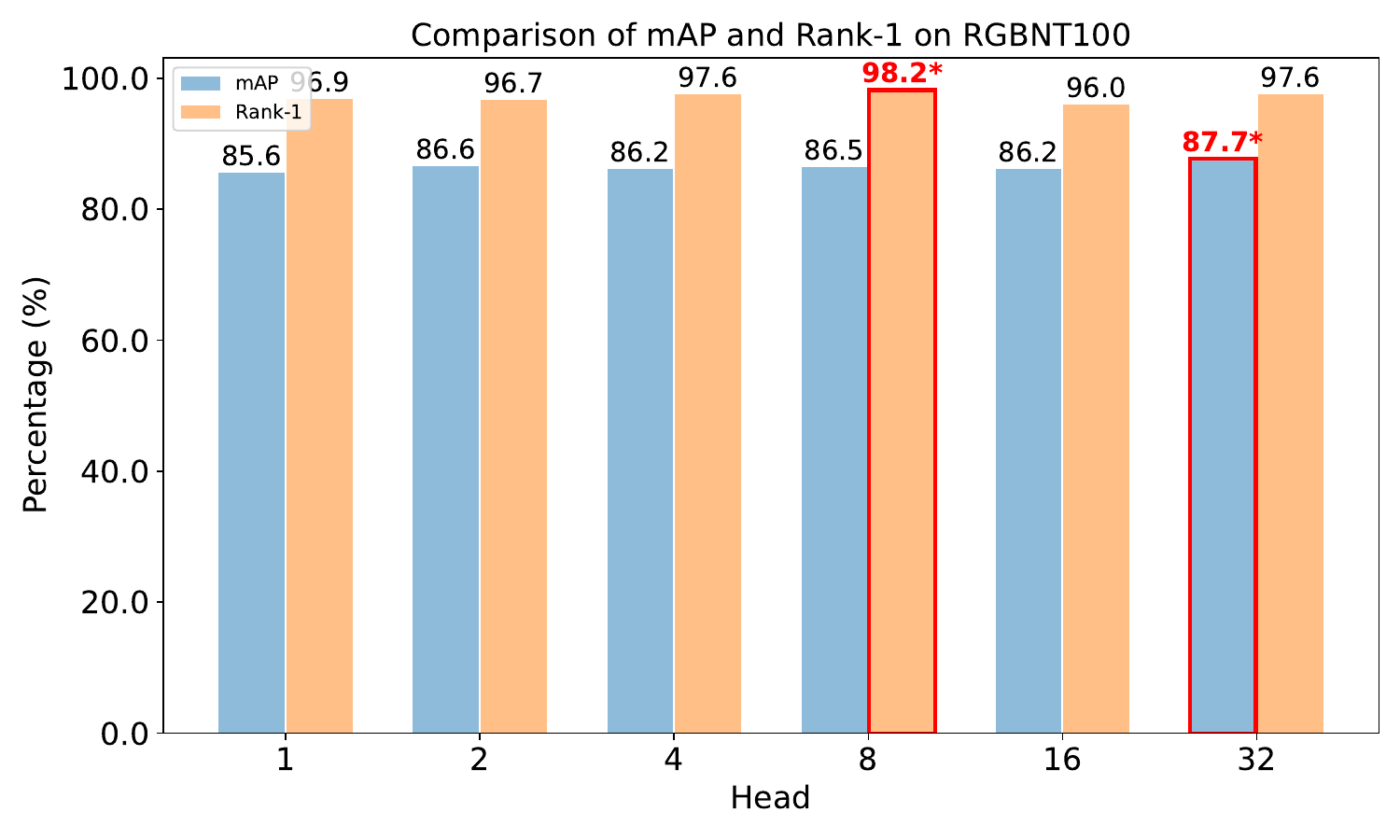}
  \vspace{-6mm}
  \caption{Comparison with different heads on RGBNT100.}
  \label{fig:heads_100}
  \vspace{-2mm}
\end{figure}
\begin{figure}[t]
  \centering
  \includegraphics[width=0.474\textwidth]{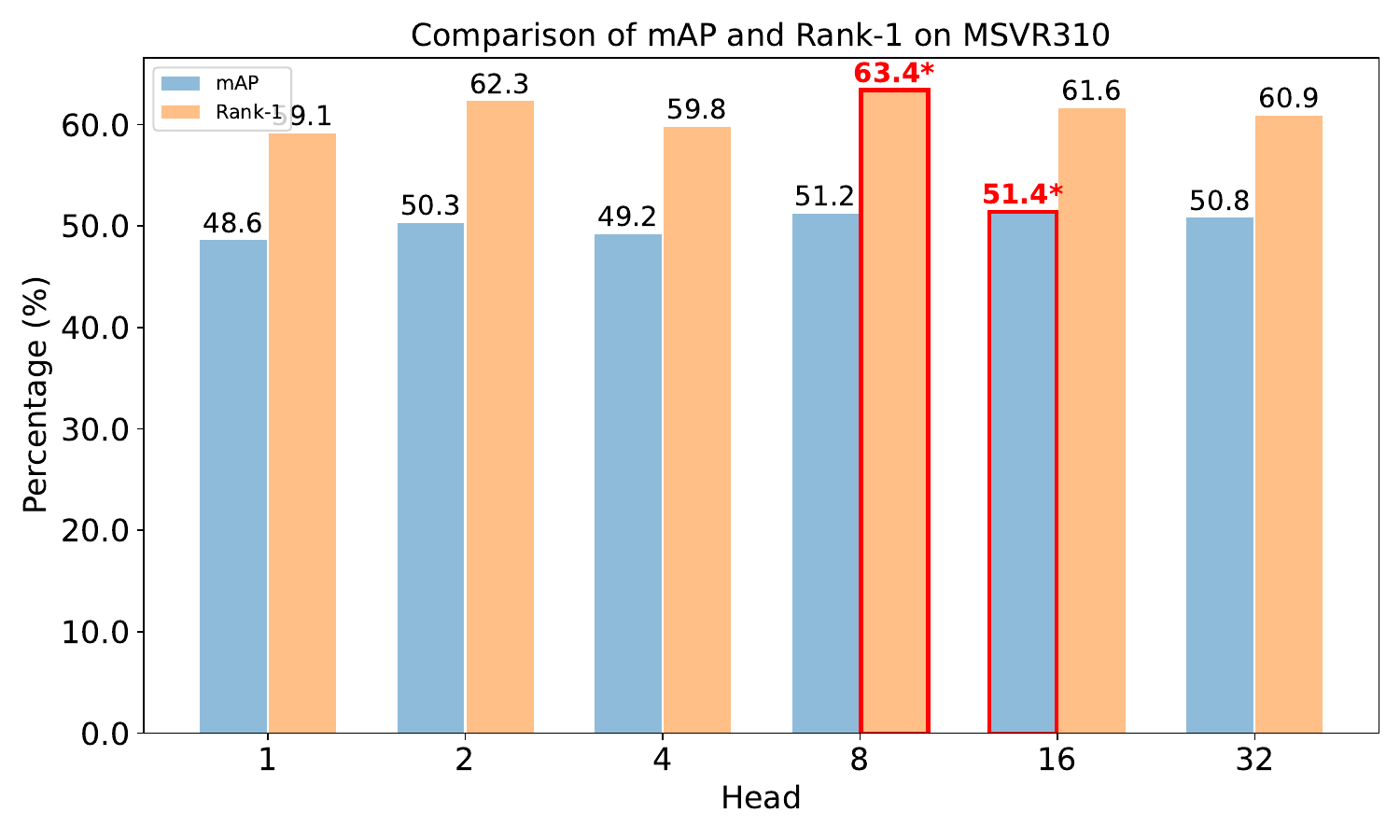}
  \vspace{-6mm}
  \caption{Comparison with different heads on MSVR310.}
  \label{fig:heads_310}
  \vspace{-2mm}
\end{figure}
\begin{table}[t]
    \vspace{-0mm}
    \centering
    \renewcommand\arraystretch{1.2}
    \setlength\tabcolsep{1 pt}
    \resizebox{0.47\textwidth}{!}
    {
    \begin{tabular}{ccccccc}
      \noalign{\hrule height 1pt}
    {\multirow{2}{*}{Methods}} &{\multirow{2}{*}{Params(M)}}&{\multirow{2}{*}{Pretrained}}&  \multicolumn{4}{c}{RGBNT201} \\\cline{4-7}
    &&&mAP&Rank-1 &Rank-5 &Rank-10 \\ \hline
    TOP-ReID & \multirow{1}{*}{324.53}&  CLIP &\underline{73.3} &\underline{77.2} &\underline{85.9} &\underline{90.1}  \\
    \rowcolor[gray]{0.92}
    \textbf{DeMo} &\textbf{\multirow{1}{*}{98.79}} & \textbf{CLIP} &\textbf{79.0} &\textbf{82.3} &\textbf{88.8} &\textbf{92.0}  \\
    \noalign{\hrule height 1pt}
    \end{tabular}
    }
    \caption{Comparison with CLIP-based TOP-ReID.}
    \label{tab:pre}
  \end{table}
\\
\textbf{Effect of Multi-Head Machanism in ATMoE.}
Fig.~\ref{fig:heads_100} and \ref{fig:heads_310} illustrate the performance of different head configurations in ATMoE on the RGBNT100 and MSVR310 datasets, respectively.
The results demonstrate a clear trend: as the number of heads increases, the model's performance improves, validating the effectiveness of the multi-head mechanism in ATMoE.
Notably, the most significant performance gain occurs when the number of heads is increased from 1 to 2.
For the RGBNT100 dataset, the model achieves the highest mAP of 87.7\% with 32 heads and the best Rank-1 accuracy of 98.2\% with 8 heads.
For the MSVR310 dataset, the model reaches its peak performance with an mAP of 51.4\% using 16 heads and a Rank-1 accuracy of 63.4\% with 8 heads.
However, to maintain consistency across datasets, particularly with RGBNT201, we standardized on 4 heads for all experiments.
The experimental results underscore that with an adequate number of heads, the model consistently achieves better performance.
These findings suggest that our DeMo can better leverage CLIP's knowledge, demonstrating the effectiveness of our modules.
\\
\textbf{Comparison with CLIP-based TOP-ReID.}
In Tab.~\ref{tab:pre}, we compare the performance of our DeMo model with the CLIP-based TOP-ReID.
TOP-ReID with CLIP visual backbone falls short in comparison to our DeMo model.
Remarkably, despite having 30.44\% parameters compared to TOP-ReID, our DeMo model achieves a higher mAP of 79.0\% and Rank-1 accuracy of 82.3\% on the RGBNT201 dataset.
These results demonstrate the effectiveness of our proposed modules in enhancing feature robustness and improving multi-modal ReID performance.
\begin{table*}[t]
    \centering
    \renewcommand\arraystretch{1.2}
    \setlength\tabcolsep{1.5pt}
    \resizebox{0.80\textwidth}{!}
    {
    \begin{tabular}{ccccccccc}
      \noalign{\hrule height 1pt}
      \multicolumn{1}{c}{\multirow{2}{*}{Methods}} &\multicolumn{1}{c}{\multirow{2}{*}{Params(M)}} &  \multicolumn{2}{c}{RGBNT201} &  \multicolumn{2}{c}{RGBNT100} & \multicolumn{2}{c}{MSVR310} \\
      \cline{3-8}
      & & mAP & Rank-1& mAP & Rank-1 & mAP & Rank-1 \\
      \hline
    MUDeep~\cite{qian2017multi} &77.75& 23.8 & 19.7 & - & - & - & - \\
    HACNN~\cite{li2018harmonious} &10.50& 21.3 & 19.0 & - & - & - & -\\
    MLFN~\cite{chang2018multi}&95.57 & 26.1 & 24.2 & - & - & - & - \\
    CAL~\cite{rao2021counterfactual} &97.62& 27.6 & 24.3 & - & - & - & - \\
    PCB~\cite{sun2018beyond}  &72.33& 32.8 & 28.1 & 57.2 &83.5 &23.2 &42.9 \\
    OSNet~\cite{zhou2019omni} &7.02& 25.4 & 22.3 & 75.0 &95.6 &28.7 &44.8 \\
    HAMNet~\cite{li2020multi} &  78.00 &27.7 &26.3 &74.5 &93.3 &27.1 &42.3\\
    CCNet~\cite{zheng2022multi} &  74.60 &- &- & 77.2 &96.3 &36.4 &\underline{55.2}\\
    IEEE~\cite{wang2022interact} & 109.22 &49.5&48.4 &-&-&-&-\\
    GAFNet~\cite{guo2022generative} &  130.00 &- &- &74.4 &93.4 &- &-\\
    LRMM~\cite{wu2025lrmm} &  86.40 &52.3 &51.1 &78.6 &\underline{96.7} &36.7 &49.7\\
    \hline
    TransReID$^*$~\cite{he2021transreid} & 278.23 &- &- &75.6 &92.9 &18.4&29.6 \\
    UniCat$^*$~\cite{crawford2023unicat}  & 259.02 &57.0 &55.7&79.4&96.2&- &- \\
    GraFT$^*$~\cite{yin2023graft} & 101.00 &- &- &76.6&94.3&-&-\\
    TOP-ReID$^*$~\cite{wang2023top} & 324.53 &\underline{72.3} &\underline{76.6} &81.2&96.4&35.9&44.6\\
    EDITOR$^*$~\cite{zhang2024magic} &118.55 & 66.5       & 68.3& 82.1 & {96.4} &39.0 & 49.3\\
    RSCNet$^*$~\cite{yu2024representation} &  124.10 & 68.2 & 72.5 &\underline{82.3} &{96.6} &\underline{39.5} &49.6\\
    \hline
    \rowcolor[gray]{0.92}
    \textbf{DeMo}$\dagger$ & 98.79 &\textbf{79.0}&\textbf{82.3}&\textbf{86.2} &\textbf{97.6} & \textbf{49.2} &\textbf{59.8}\\
    \noalign{\hrule height 1pt}
    \end{tabular}
    }
    \caption{Performance comparison in our framework.
    The best and second results are in bold and underlined, respectively.
    The symbol $\dagger$ denotes CLIP-based methods, $*$ indicates ViT-based methods and others are CNN-based methods.}
    \label{tab:params}
    \vspace{-1mm}
    \end{table*}
  \begin{table}[th]
    \centering
    \renewcommand\arraystretch{1.2}
    \setlength\tabcolsep{4.5pt}
    \resizebox{0.47\textwidth}{!}
    {
    \begin{tabular}{cccccccc}
        \noalign{\hrule height 1pt}
        \multicolumn{1}{c}{\multirow{2}{*}{\textbf{Index}}}                   &\multicolumn{3}{c}{\textbf{Modules}}                                  & \multicolumn{2}{c}{\textbf{RGBNT100}} & \textbf{Params}                 & \textbf{FLOPs}                  \\
        \cmidrule(r){2-4} \cmidrule(r){5-6} \cmidrule(r){7-7} \cmidrule(r){8-8}
   & \textbf{PIFE}              & \textbf{HDM}                & \textbf{ATMoE}                   & \textbf{mAP}    & \textbf{R-1}   & \textbf{M}                      & \textbf{G}                      \\ \hline
    A                  & \ding{53}                  & \ding{53}                  & \ding{53}                    & 82.5  & 95.0 & 86.22                  & 34.28                  \\
    B                  & \ding{51}                  & \ding{53}                  & \ding{53}                      & 83.1  & 96.4 & 87.81                  & 34.28                  \\
    \multirow{1}{*}{C} & \multirow{1}{*}{\ding{51}} & \multirow{1}{*}{\ding{51}} & \multirow{1}{*}{\ding{53}}    & 85.4  & 96.8 & 95.35                  & 35.09 \\
    \rowcolor[gray]{0.92}
    \multirow{1}{*}{D} & \multirow{1}{*}{\ding{51}} & \multirow{1}{*}{\ding{51}} & \multirow{1}{*}{\ding{51}}   &\textbf{86.2} 	 &\textbf{97.6}  & 98.18                 & 35.10 \\
    \noalign{\hrule height 1pt}
    \end{tabular}
    }
    \caption{Performance comparison with different modules.}
    \label{tab:ablation_vehicle}
    \vspace{-6mm}
  \end{table}
  \begin{figure}[t]
    \centering
    \includegraphics[width=0.444\textwidth]{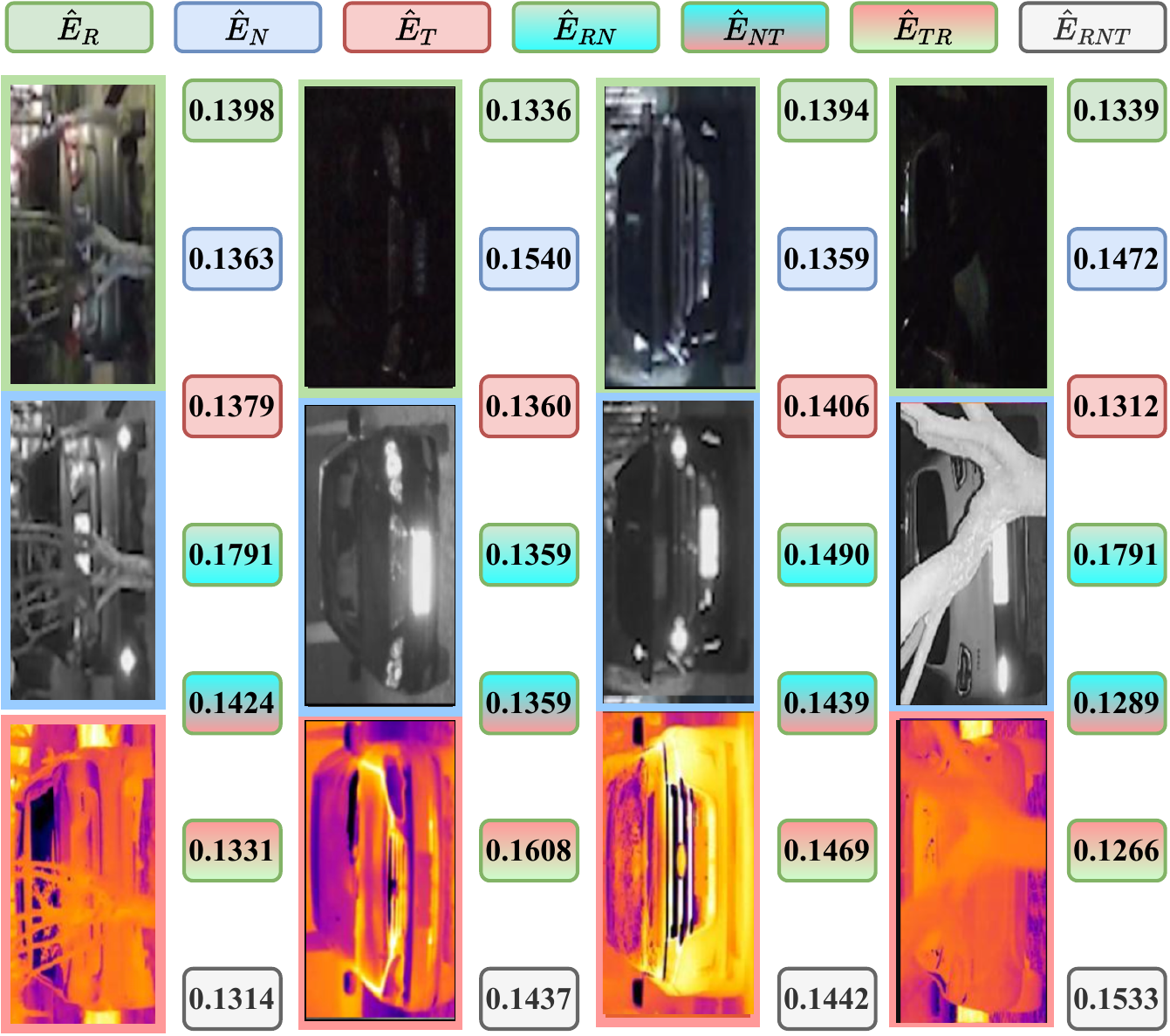}
    \caption{Dynamic weights in ATMoE on RGBNT100.}
    \label{fig:dyn_weight}
    \vspace{-5mm}
  \end{figure}
  \begin{figure*}[t]
    \centering
    \includegraphics[width=0.76\textwidth]{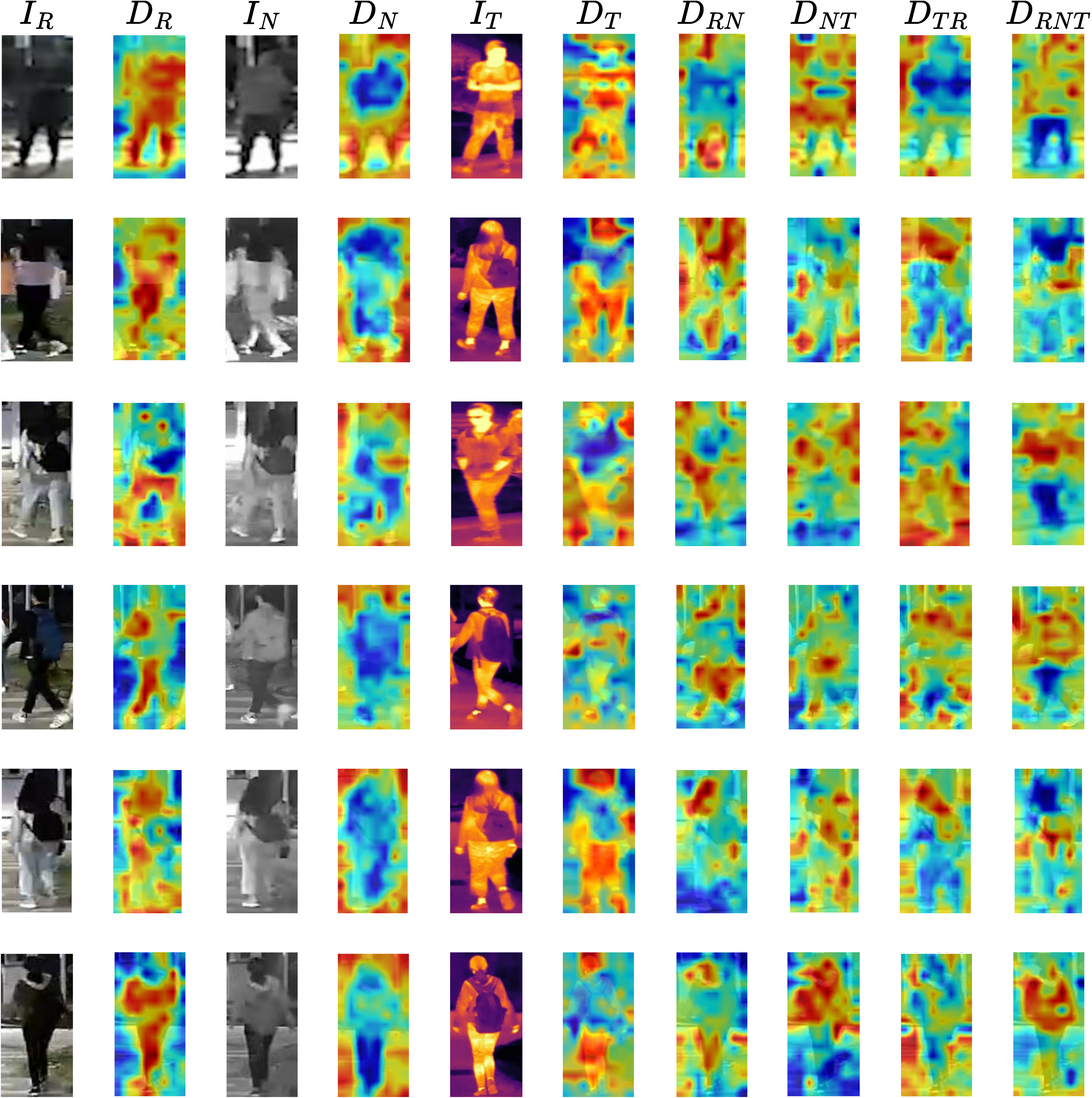}
    \caption{Activation maps of decouple features in HDM.
    $I_{m}, m \in \{R, N, T\}$ denotes the RGB, NIR and TIR input images, respectively.
    $D_{m'}, m' \in \{R, N, T, RN, NT, TR, RNT\}$ denotes the different decoupled features in HDM.}
    \label{fig:person-gradcam}
  \end{figure*}
  \begin{figure*}[t]
    \centering
    \includegraphics[width=1\textwidth]{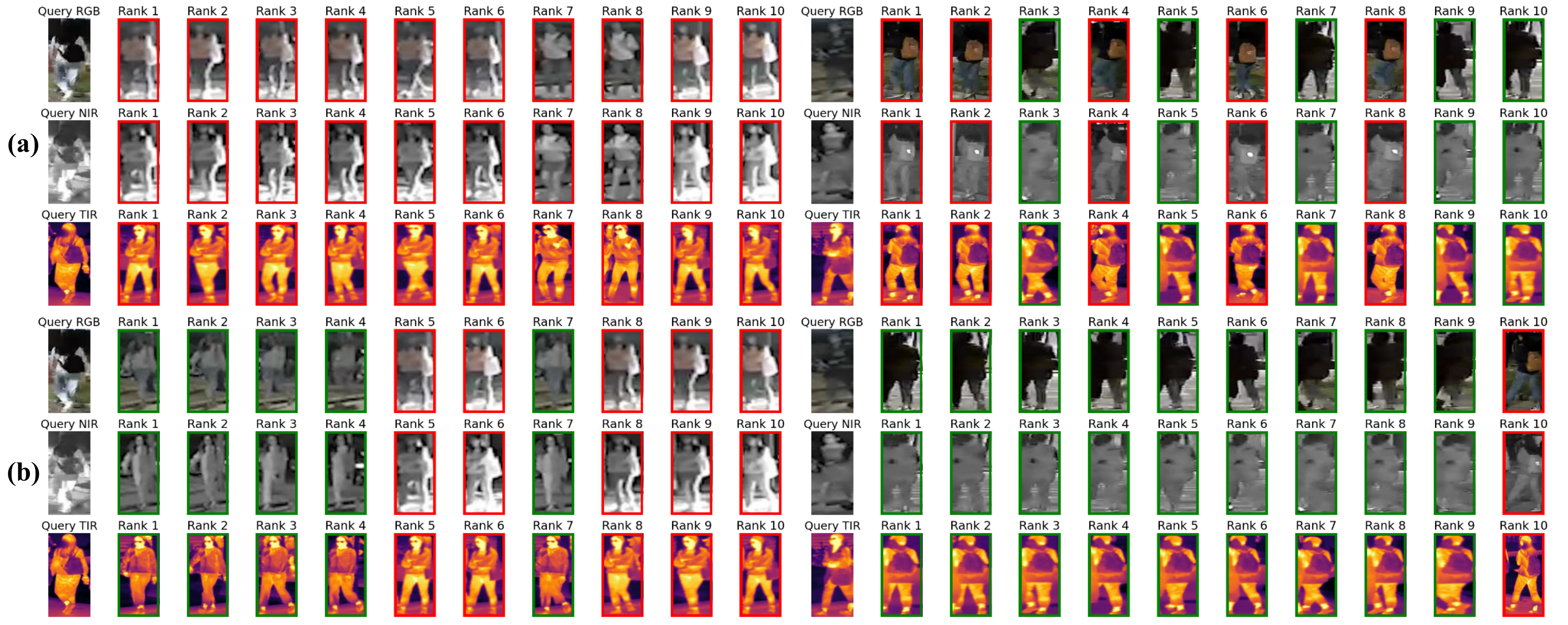}
    \caption{Rank list comparison with different methods.
    (a) TOP-ReID.
    (b) Our proposed DeMo.
    }
    \label{fig:rank_topreid}
  \end{figure*}
  \begin{figure*}[t]
    \centering
    \includegraphics[width=1\textwidth]{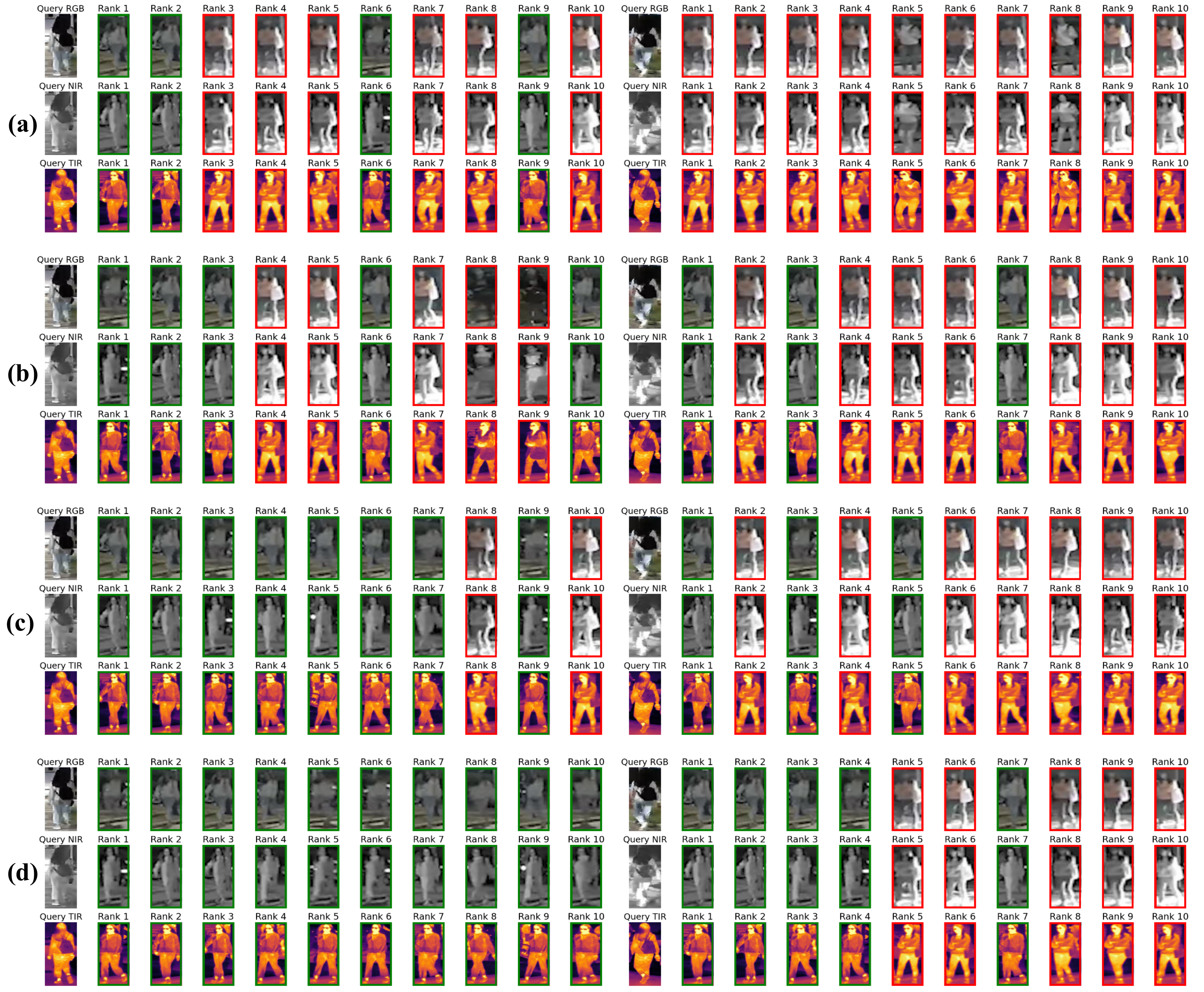}
    \vspace{-4mm}
    \caption{Rank list comparison with different modules.
    (a) Baseline.
    (b) Baseline + PIFE.
    (c) Baseline + PIFE + HDM.
    (d) Baseline + PIFE + HDM + ATMoE.
    The green box indicates the correct match, while the red box indicates the incorrect match.}
    \label{fig:rank_modules}
    \vspace{-4mm}
  \end{figure*}
\subsection{Parameter Comparisons with Other Methods}
In Tab.~\ref{tab:params}, we compare the trainable parameters of our model with other state-of-the-art methods.
Generally, CNN-based methods have fewer parameters compared to Transformer-based methods with comparable performance.
Among the Transformer-based methods, our DeMo model achieves competitive performance with fewer parameters.
Specifically, DeMo uses only 30.44\% of the parameters of TOP-ReID, yet it achieves an average 8.33\% mAP improvement across three datasets, highlighting the efficiency and effectiveness of our model.
\subsection{Generalization to Vehicle Datasets}
In Tab.~\ref{tab:ablation_vehicle}, we evaluate the effectiveness of key components using the RGBNT100 vehicle dataset.
Comparing Model A with Model B, we observe a substantial improvement in mAP, highlighting the critical role of integrating the detailed information from the local patches across modalities.
Model C introduces the HDM, which explicitly decouples features and further enhances performance.
Additionally, Model D integrates the ATMoE, leading to a significant performance boost through the use of multi-head attention-guided experts.
Notably, this improvement is achieved while maintaining FLOPs and parameter counts at a reasonable level, underscoring the overall effectiveness of our DeMo model.
\section{C. Visualization Results}
\textbf{Dynamic Weight Visualizations on Vehicle Dataset.}
In Fig.\ref{fig:dyn_weight}, we illustrate the dynamic weights assigned to different instances on the RGBNT100 dataset.
The weights of the various decoupled features vary across instances, demonstrating ATMoE's ability to adjust feature importance based on the specific characteristics of each instance.
As observed in the visualizations, modalities containing more detailed information receive greater emphasis.
It is worth noting that the dynamic weights are averaged across all heads in ATMoE, resulting a relatively smooth distribution.
However, the weights can vary significantly across different heads, reflecting the diverse attention patterns learned by the model.
\\
\textbf{Activation Maps of Decoupled Features in HDM.}
Fig.\ref{fig:person-gradcam} displays the activation maps of decoupled features within HDM.
These activation maps correspond to the RGB, NIR, and TIR input images, each highlighting the decoupled features.
Overall, the activation maps reveal that the decoupled features focus on distinct discriminative regions of the input images, illustrating the effectiveness of the decoupling process in HDM with more diverse and informative features.
\\
\textbf{Rank List Comparison with TOP-ReID.}
Fig.~\ref{fig:rank_topreid} presents a comparison of rank lists between TOP-ReID and our DeMo.
We specifically select challenging examples from the RGBNT201 dataset for this comparison.
Our DeMo model consistently outperforms TOP-ReID, with fewer incorrect matches and more accurate results.
\\
\textbf{Rank List Comparison with Different Modules.}
In Fig.~\ref{fig:rank_modules}, we compare the rank lists across various modules.
The baseline model exhibits subpar performance, with numerous incorrect matches.
Incorporating PIFE results in a modest improvement, though the model continues to encounter significant issues with incorrect matches.
The addition of HDM further enhances performance, reducing the number of incorrect matches.
Finally, the inclusion of ATMoE leads to the best performance, with the fewest incorrect matches overall, verifying the effectiveness of our modules.
\end{document}